\documentclass[10pt]{article}

\usepackage{times}
\usepackage{tiberio}

\usepackage[pagebackref=true,breaklinks=true,letterpaper=true,colorlinks,bookmarks=false]{hyperref}

\begin{document}

\title{Scalable Inference for Latent Dirichlet Allocation}
\author{James Petterson, Tib\'erio Caetano\thanks{NICTA's Statistical Machine Learning program, Locked Bag 8001, ACT 2601, Australia, and Research School of Information Sciences and Engineering, Australian National University, ACT 0200, Australia. NICTA is funded by the Australian Government'Õs Backing Australia'Õs Ability initiative, and the Australian Research CouncilÕ's ICT Centre of Excellence program. e-mails: \texttt{first.last@nicta.com.au}}}

\maketitle

\begin{abstract}
We investigate the problem of learning a topic model -- the well-known Latent Dirichlet Allocation -- in a distributed manner, using a cluster of C processors and dividing the corpus to be learned equally among them. We propose a simple approximated method that can be tuned, trading speed for accuracy according to the task at hand. Our approach is asynchronous, and therefore suitable for clusters of heterogenous machines. 

\end{abstract}

\section{Introduction}

Very large datasets are becoming increasingly common -- from specific collections, such as Reuters and PubMed, to very broad and large ones, such as the images and metadata of sites like Flickr, scanned books of sites like Google Books and the whole internet content itself.
Topic models, such as Latent Dirichlet Allocation (LDA), have proved to be a useful tool to model such collections, but suffer from scalability limitations. Even though there has been some recent advances in speeding up inference for such models, this still remains a fundamental open problem.

\section{Latent Dirichlet Allocation}

\begin{figure}[tb]
\centerline{\includegraphics[width=0.8\textwidth]{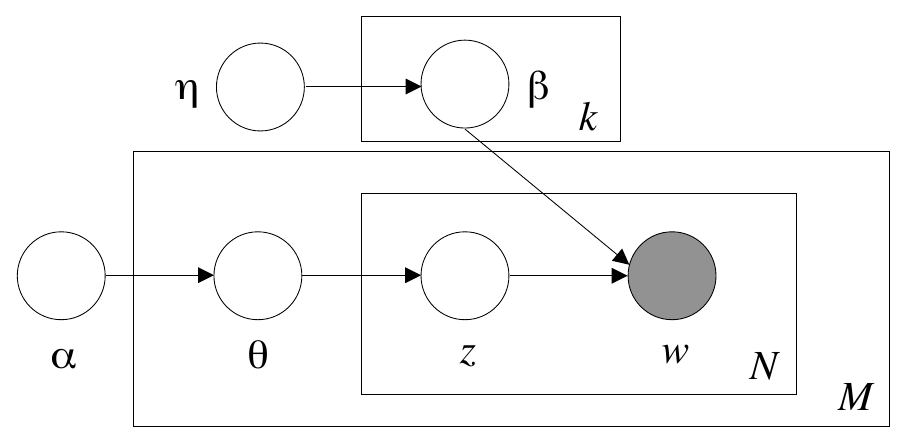}}
\caption{ LDA model. }
\label{fig:lda}
\end{figure}

Before introducing our method we briefly describe the Latent Dirichlet Allocation (LDA) topic model \cite{BleNgJor03}.
In LDA (see Figure \ref{fig:lda}), each document is modeled as a mixture over $K$ topics, and each topic has a multinomial distribution $\beta_k$ over a vocabulary of $V$ words (please refer to table \ref{tab:notation} for a summary of the notation used throughout this paper).
For a given document $m$ we first draw a topic distribution $\theta_{m}$ from a Dirichlet distribution parametrized by $\alpha$. Then, for each word $n$ in the document we draw a topic $z_{m,n}$ from a multinomial distribution with parameter $\theta_{m}$. Finally, we draw the word $n$ from the multinomial distribution parametrized by $\beta_{z_{m,n}}$.

\begin{table}[h]
\caption{Notation}
\label{tab:notation}
\centering
\begin{tabular}{ll}
variable & description\\
\hline
$D_{train}$ & training document corpus\\
$D_{test}$ & testing document corpus\\
$K$ & number of topics\\
$M$ & number of documents\\
$N_m$ & number of words in document m\\
$V$ & dictionary size\\
$C$ & number of CPUs\\
$\alpha$ & Dirichlet prior for $\theta$ (hyperparameter) \\
$\eta$ & Dirichlet prior for $\beta$ (hyperparameter) \\
$\theta$ & distribution of topics per document\\
$\beta$ & distribution of topics per word\\
$z_{m,n}$ & topic $(1..K)$ of word $n$ of document $m$\\
$w_{m,n}$ & term index $(1..V)$ of word $n$ of document $m$\\
$n_{k,v}$ & number of times the term $v$ has been observed with topic $k$\\
$n_{k,v}^l$ & local modifications to $n_{k,v}$\\
$n_k$ & number of times topic $k$ has been observed in all documents\\
$n_{m,k}$ & number of times topic $k$ has been observed in a word of document $m$\\
$n_m$ & number of words in document $m$\\
\end{tabular}
\end{table}

\subsection{Inference in LDA}

Many inference algorithms for LDA have been proposed, such as variational Bayesian (VB) inference \cite{BleNgJor03}, expectation propagation (EP) \cite{MinLaf02}, collapsed Gibbs sampling \cite{GriSte04, Hei04} and collapsed variational Bayesian (CVB) inference \cite{TehNewWel06}. In this paper we will focus on collapsed Gibbs sampling.

\subsection{Collapsed Gibbs sampling}

Collapsed Gibbs sampling is an MCMC method that works by iterating over each of the latent topic variables $z_1$, ..., $z_n$, sampling each $z_i$ from $P(z_i | z_{\neg i})$. This is done by integrating out the other latent variables ($\theta$ and $\beta$). We are not going to dwell on the details here, since this has already been well explained in \cite{GriSte04, Hei04}, but in essence what we need to do is to sample from this distribution:

\begin{align}
\label{eq:z_sampling}
p(z_i=k | z_{\neg i}, w) & \propto   \frac { \left( n_{k,v,\neg i} + \eta \right)   }{ \sum_{v=1}^V \left( n_{k,v,\neg i} + \eta \right) } \left( n_{m,k,\neg i} + \alpha \right)\\
& \propto   \frac { \left( n_{k,v,\neg i} + \eta \right)   }{ \left( n_{k,\neg i} + V \eta \right) } \left( n_{m,k,\neg i} + \alpha \right)
\end{align}

In simple terms, to sample the topic of a word of a document given all the other words and topics we need, for each $k$ in $\{1,\dots,K\}$:
\begin{enumerate}
\item $n_{k,v,\neg i}$: the total number of times the word's term has been observed with topic $k$ (excluding the word we are sampling for).
\item $n_{k, \neg i}$: the total number of times topic $k$ has been observed in all documents (excluding the word we are sampling for).
\item $n_{m,k,\neg i}$: the number of times topic $k$ has been observed in a word of this document (excluding the word we are sampling for).
\end{enumerate}

\section{Related work}

There has been research in different approaches to increase the efficiency and/or scalability of LDA. We are going to discuss them next.

\subsection{Faster sampling}

The usual approach to draw samples of $z$ using \eq{eq:z_sampling} is to compute a normalization constant $Z = \sum_{k=1}^K p(z_i=k | z_{\neg i}, w)$ to obtain a probabily distribution that can be sampled from:
\begin{align}
\label{eq:z_sampling_norm}
p(z_{i=k} | z_{\neg i}, w) = \frac{1}{Z} \frac { \left( n_{k,v,\neg i} + \eta \right)   }{ \left( n_{k,\neg i} + V \eta \right) } \left( n_{m,k,\neg i} + \alpha \right)
\end{align}

This leads to a complexity for each iteration of standard Gibbs sampling of $O(N_T K)$, where $N_T$ is the total number of words in the corpus, and $K$ is the number of topics. 

\cite{PorNewIhlAsuetal08} proposed a way to avoid computing \eq{eq:z_sampling} for each $K$ by getting an upper bound on $Z$ using Holder's inequality and computing \eq{eq:z_sampling} for the most probable topics first, leading to a speed up of up to 8x of the sampling process.

\cite{YaoMimMcc09} broke \eq{eq:z_sampling} in three components and took leverage on the resulting sparsity in $k$ of some of them -- that, combined with an efficient storage scheme led to a speed up of the order of 20x.

\subsection{Parallelism}

A complementary approach for scalability is to share the processing among several CPUs/cores, in the same computer (multi core) or in different computers (clusters).

\subsubsection{Fine grained parallelism}

In most CPU architectures the cost incurred in creating \emph{threads}/\emph{processes} and synchronizing data among them can be very significant, making it infeasible to share a task in a fine-grained manner. One exception, however, are Graphics Processing Units (GPUs). Since they were originally designed to parallelize jobs in the pixel level, they are well suited for fine-grained parallelization tasks.

\cite{MasHamShiOgu09} proposed to use GPUs to parallelize the sampling at the topic level. Although their work was with collapsed variational Bayesian (CVB) inference \cite{TehNewWel06}, it could probably be extended to collapsed Gibbs sampling. It's interesting to note that this kind of parallelization is complementary to the document-level one (see next section), so both can be applied in conjunction.

\subsubsection{Coarse grained parallelism}

Most of the work on parallelism has been on the document level -- each CPU/core is responsible for a set of documents.

Looking at equation \eq{eq:z_sampling} it can be seen that in the right hand side we have a document specific variable ($n_{m,k}$). Only $n_{k,v}$ (and its sum, $n_k$), on the left hand side, is shared among all documents. Using this fact, \cite{NewAsuSmyWel07} proposed to simply compute a subset of the documents in each CPU, synchronizing the global counts ($n_{k,v}$) at the end of each step. This is an approximation, since we are no longer sampling from the true distribution, but from a noisy version of it. They showed, however, that it works well in practice. They also proposed a more principled way of sharing the task using a hierarchical model and, even though that was more costly, the results were similar.

\cite{AsuSmyWel08} proposed a similar idea, but with an asynchronous model, where there is no global synchronization step (as there is in \cite{NewAsuSmyWel07}).

\section{Our method}

We follow \cite{AsuSmyWel08} and work in a coarse-grained asynchronous parallelism, dividing the task at the document level. For simplicity, we split the $M$ documents among the $C$ CPUs equally, so that each CPU receives $\frac{M}{C}$ documents\footnote{This is not strictly necessary: when working with a cluster of heterogeneous CPUs it might be more interesting to split proportionally to the processing power of each CPU.}. We then proceed in the usual manner, with each CPU running the standard Gibbs sampling in its set of documents. Each CPU, however, keeps a copy of all its modifications to $n_{k,v}$ and, at the end of each iteration, stores them in a file in a shared filesystem. Right after that, it reads all modifications stored by other CPUs and incorporates them to its $n_{k,v}$. This works in an asynchronous manner, with each CPU saving its modifications and reading other CPU's modifications at the end of each iteration. The algorithm is detailed in \ref{alg:simple}.

\begin{algorithm}[ht]
  \caption{Simple sharing}
  \begin{algorithmic}
    \STATE {\bfseries Input:} $\alpha$, $\eta$, $K$, $D_{train}$, $C$, $num\_iter$
    \STATE Randomly initialize $z_{m,n}$, updating $n_{k,v}$ and $n_{k,v}^l$ accordingly.
    \STATE Save $n_{k,v}^l$ to a file
    \FOR{$t = 1$ to $num\_iter$}
    \STATE Run collapsed Gibbs sampling, updating $z_{m,n}$, $n_{k,v}$ and $n_{k,v}^l$
    \STATE Save $n_{k,v}^l$ to a file
    \STATE Load modifications to $n_{k,v}$ from other CPUs
    \ENDFOR
  \end{algorithmic}
  \label{alg:simple}
\end{algorithm}

We first note that, in this simple algorithm, the complexity of the sampling step is $O(N_c K)$ (whre $N_c$ is the number of words being processed in CPU $c$), while the synchronization part takes $O(C K V)$ (we save a $K$x$V$ matrix once and load it $C-1$ times). Plugging in the following values, based on a standard large scale task:

\begin{itemize}
\item $K$ = $500$ topics
\item $C$ = $100$ CPUs
\item $N_c$ = $10^7$ words
\item $V$ = $10^5$ terms
\end{itemize}
 we get similar values for the sampling and the synchronization steps. That, however, doesn't take into account the constants. In our experiments, with these parameters a sampling step will take approximately 500 seconds, while the synchronization will take around 20,000 seconds (assuming a 1Gbit/s ethernet connection shared among all CPUs). The bottleneck is clearly in the synchronization step.
 
We propose, therefore, a variation of the first algorithm. When saving the modifications at the end of an iteration, only save those that are \emph{relevant} -- more formally, save (in a sparse format) only those items of $n_{k,v}^l$ for which

\begin{align}
\frac{n_{k,v}^l}{n_{k,v}} > threshold
\end{align}

where $threshold$ is a parameter that can range from 0 to 1. The algorithm is detailed in \ref{alg:sparse}.
Note that setting $threshold$ to zero recovers Algorithm \ref{alg:simple}.

\begin{algorithm}[ht]
  \caption{Sparse sharing}
  \begin{algorithmic}
    \STATE {\bfseries Input:} $\alpha$, $\eta$, $K$, $D_{train}$, $C$, $num\_iter$
    \STATE Randomly initialize $z_{m,n}$, updating $n_{k,v}$ and $n_{k,v}^l$ accordingly.
    \STATE Save $n_{k,v}^l$ to a file
    \FOR{$t = 1$ to $num\_iter$}
    \STATE Run collapsed Gibbs sampling, updating $z_{m,n}$, $n_{k,v}$ and $n_{k,v}^l$
    \FOR{$k = 1$ to $K$}
    \FOR{$v = 1$ to $V$}
    \STATE Save $n_{k,v}^l$ if $\frac{n_{k,v}^l}{n_{k,v}} > threshold$
    \ENDFOR
    \ENDFOR
    \STATE Load modifications to $n_{k,v}$ from other CPUs
    \ENDFOR
  \end{algorithmic}
  \label{alg:sparse}
\end{algorithm}

\section{Experiments}

\subsection{Datasets}

We ran our experiments in three datasets: NIPS full papers (books.nips.cc), Enron emails (www.cs.cmu.edu/$\sim$enron) and KOS (dailykos.com)\footnote{We used the preprocessed datasets available at \href{http://archive.ics.uci.edu/ml/datasets/Bag+of+Words}{http://archive.ics.uci.edu/ml/datasets/Bag+of+Words}.}. Each dataset was split in 90\% for training and 10\% for testing. Details on the parameters of the datasets are shown in table \ref{tab:datasets}.

All experiments were ran in a cluster of 11 machines, each one with a dual-core AMD64 2.4 GHz CPU and 8 Gb of RAM (22 CPUs total). All machines share a network file system over an 1GB Ethernet network.

\begin{table}[tbp]
\caption{Parameters of the three datasets used.}
\label{tab:datasets}
\centering
\begin{tabular}{lrrr}
& NIPS & Enron & KOS\\
\hline
number of documents in $D_{train}$    & 1350    & 35,874     & 3,087 \\
number of documents in  $D_{test}$     & 150      & 3,987       & 343 \\
total number of words  &  1,932,364  & 6,412,171 & 467,713 \\
vocabulary size $V$                & 12,419 & 28,102     & 6,906\\
\end{tabular}
\end{table}

We used a fixed set of LDA parameters: $K=50$ (unless otherwise noticed), $\alpha=0.1$, $\eta=0.01$ and 1500 iterations of the Gibbs sampler. To compare the quality of different approximations we computed the \emph{perplexity} of a held-out test set. The perplexity is commonly used in language modeling: it is equivalent to the inverse of the geometric mean per-word likelihood. Formally, given a test set of $M_{test}$ documents:

\begin{align}
\label{eq:perplexity}
perplexity(D_{test}) = exp \left\{ -\frac{\sum_{m=1}^{M_{test}} \sum_{n=1}^{N_m} log~p(w_{m,n})}{\sum_{m=1}^{M_{test}} N_m} \right\}
\end{align}

\subsection{Results}

\begin{figure*}[tbp]
\centerline{
\includegraphics[width=0.45\textwidth]{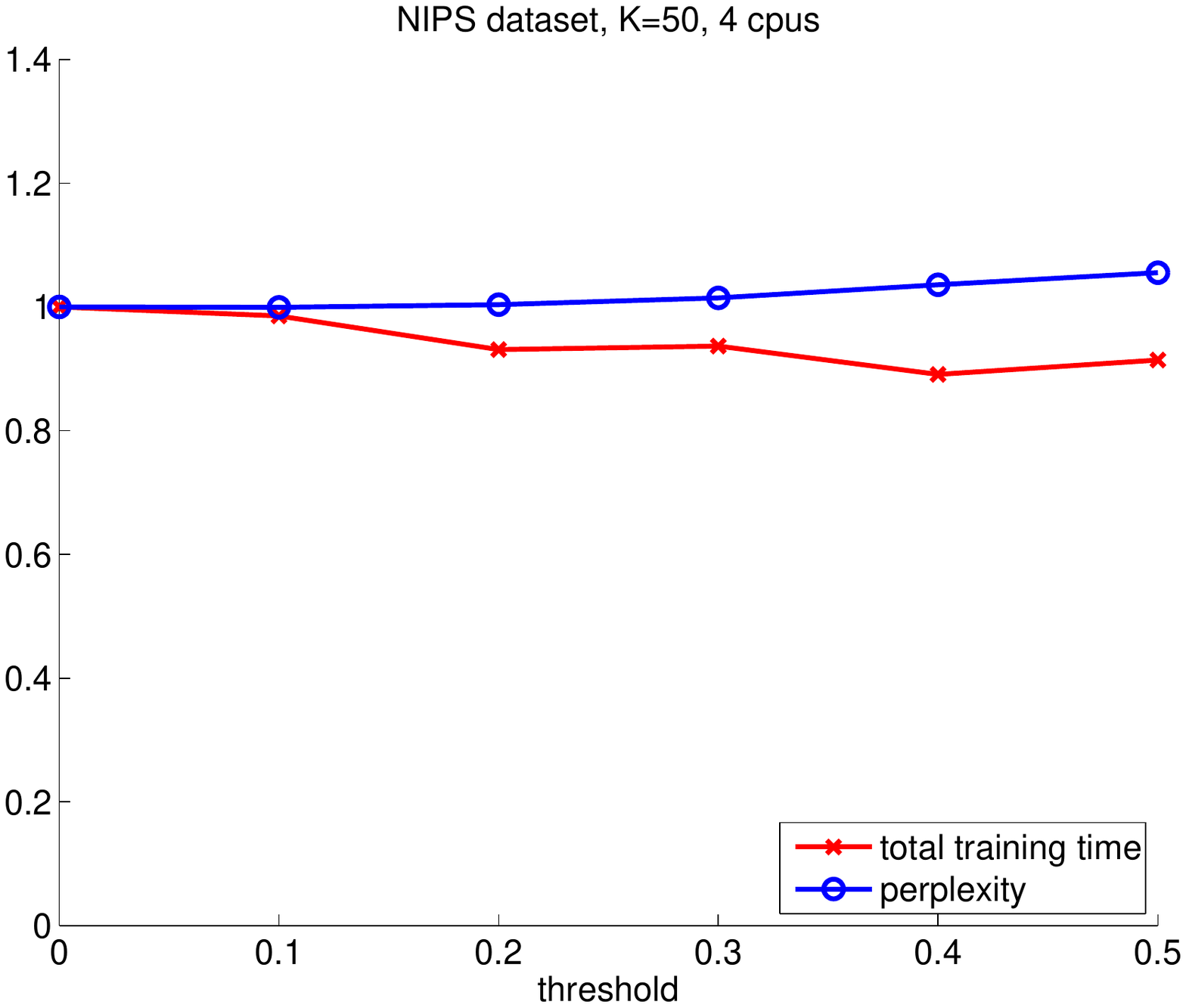}
\includegraphics[width=0.45\textwidth]{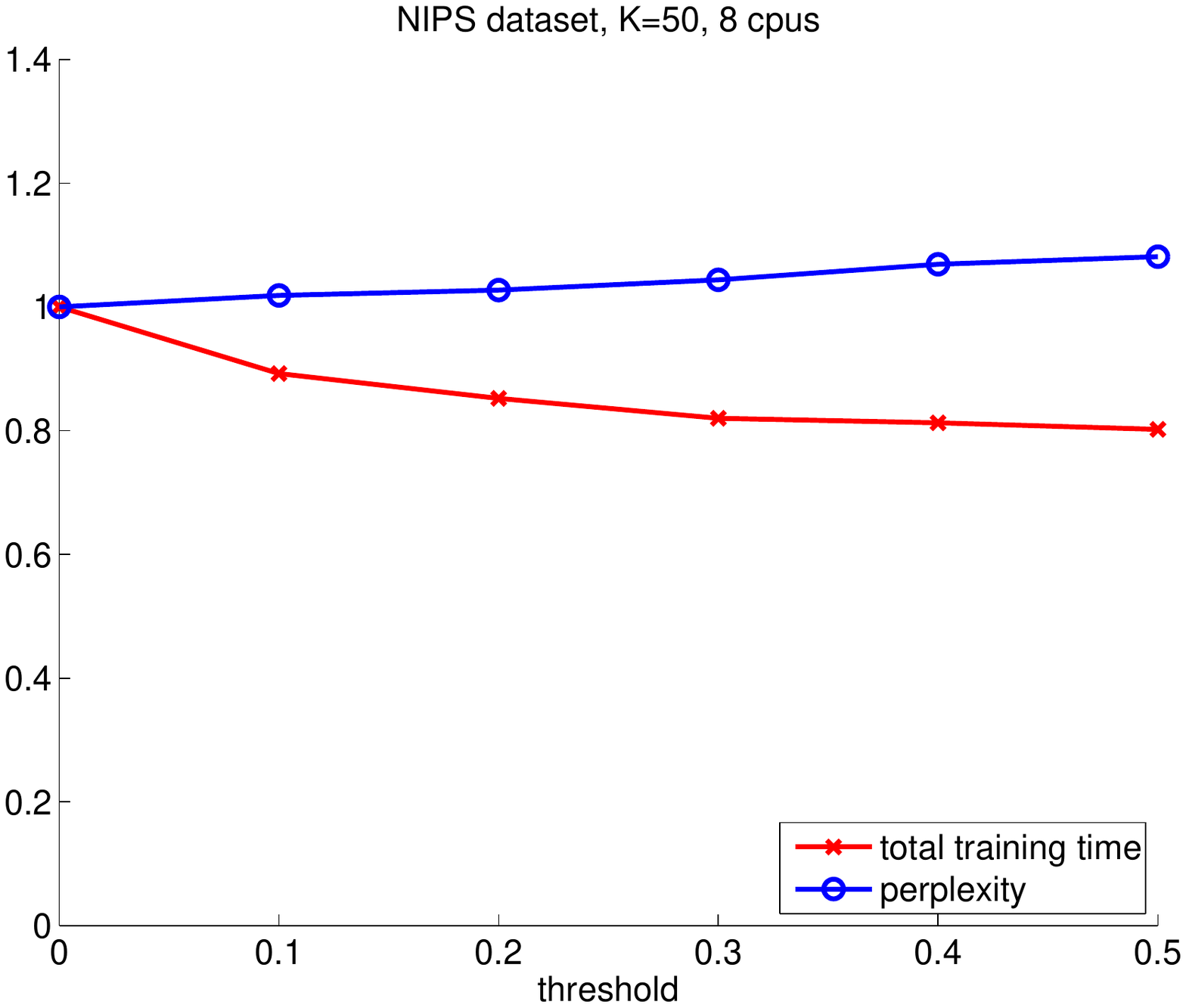}
\includegraphics[width=0.45\textwidth]{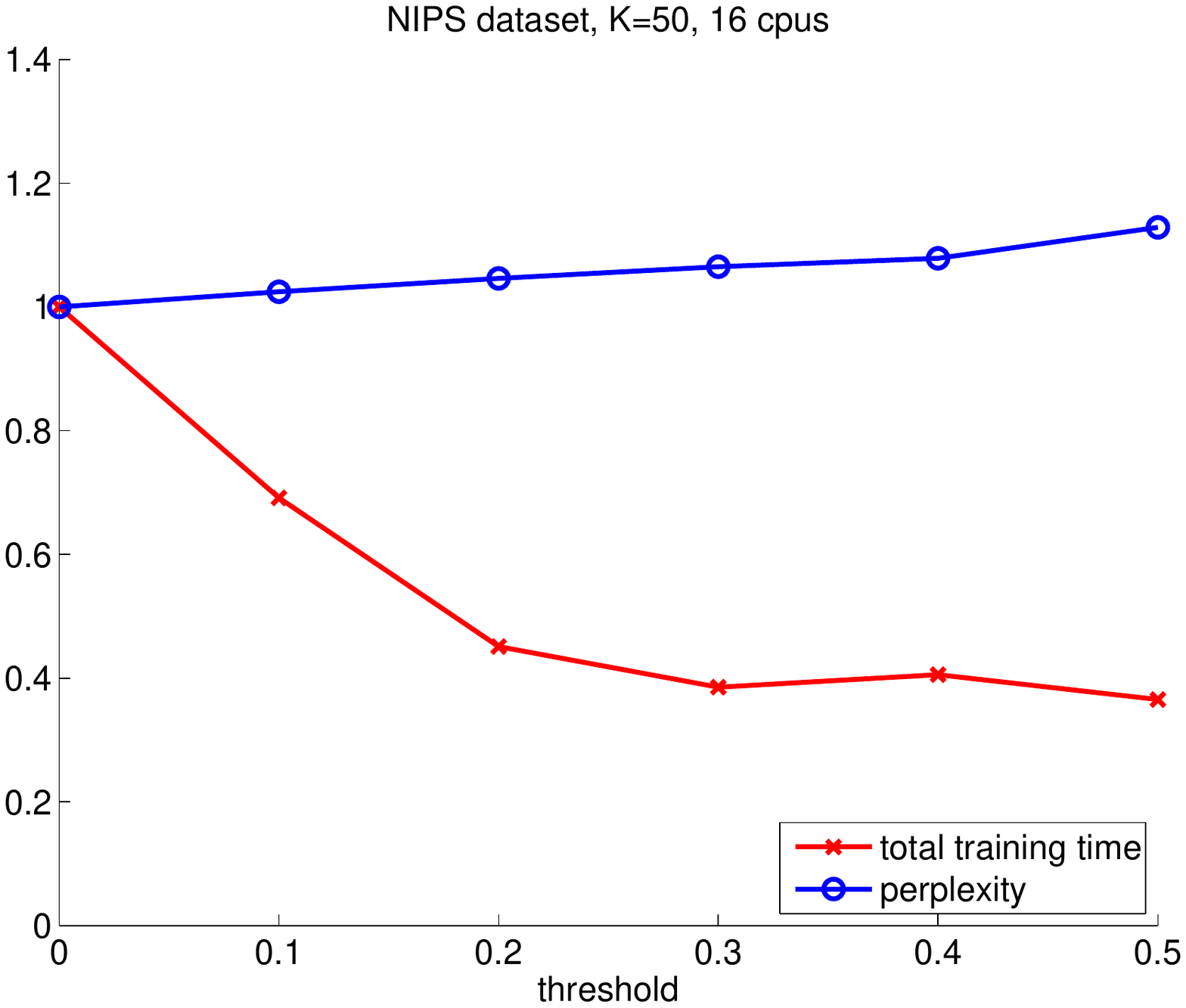}
}
\vspace{-30mm}
\centerline{
\includegraphics[width=0.45\textwidth]{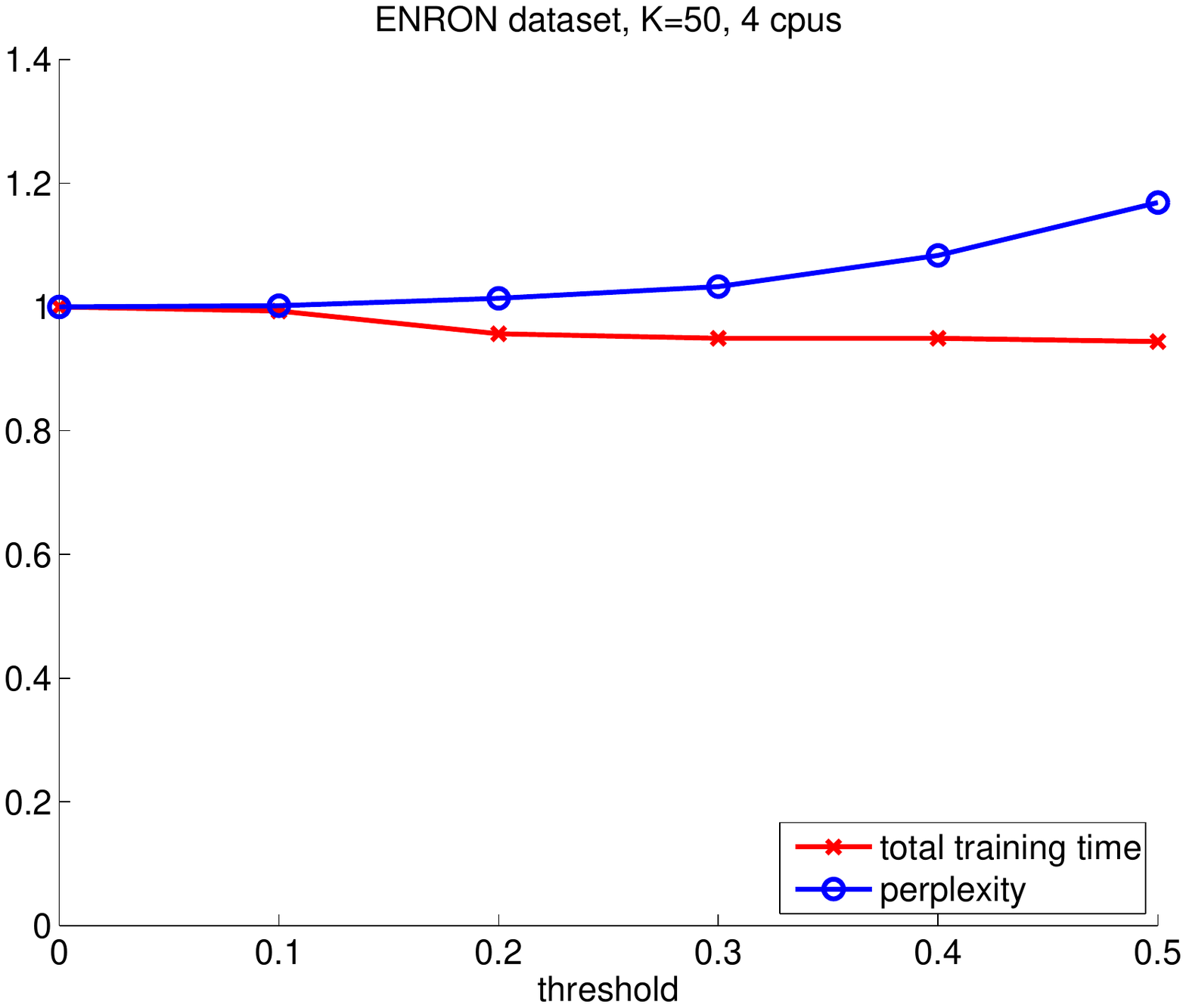}
\includegraphics[width=0.45\textwidth]{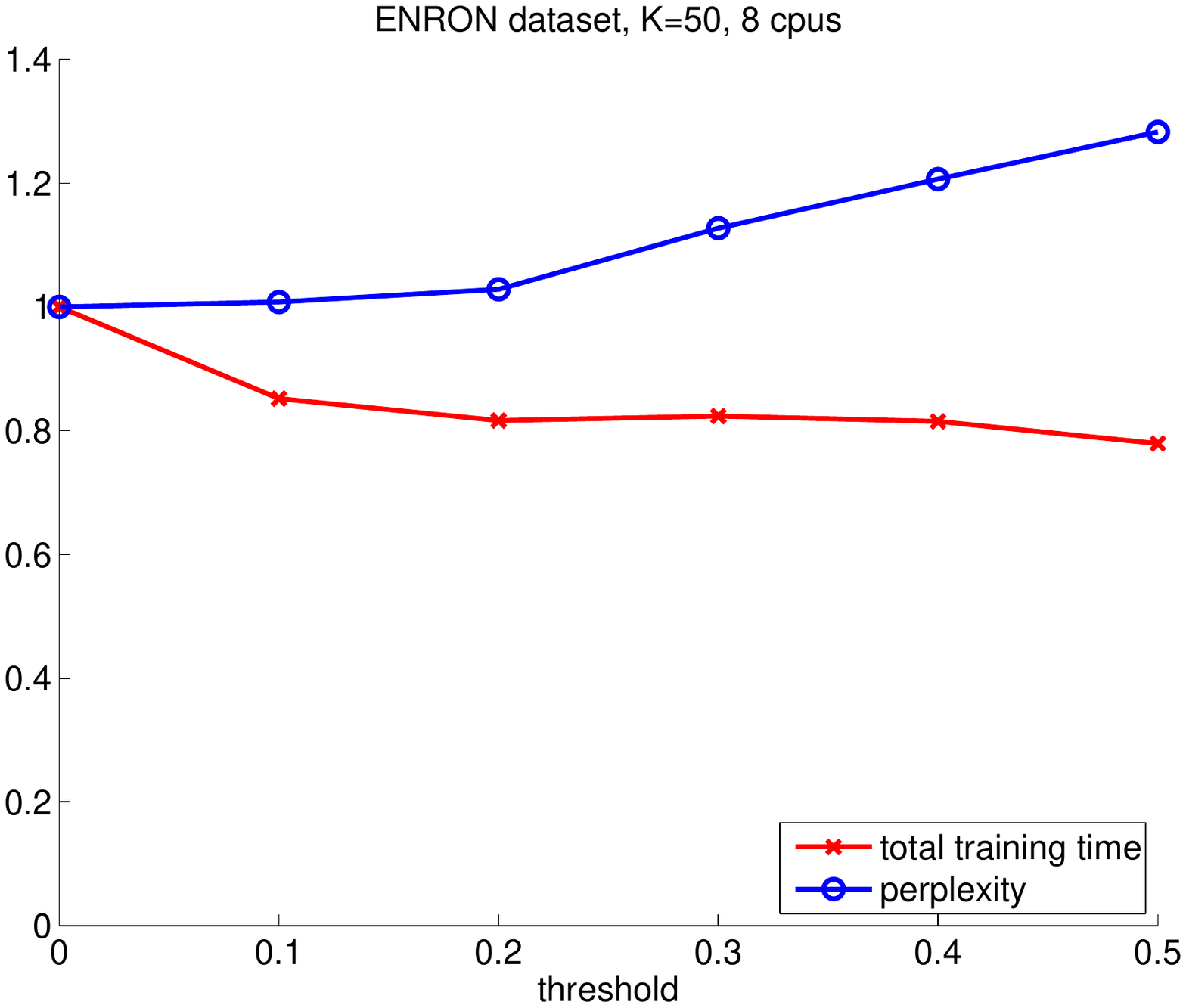}
\includegraphics[width=0.45\textwidth]{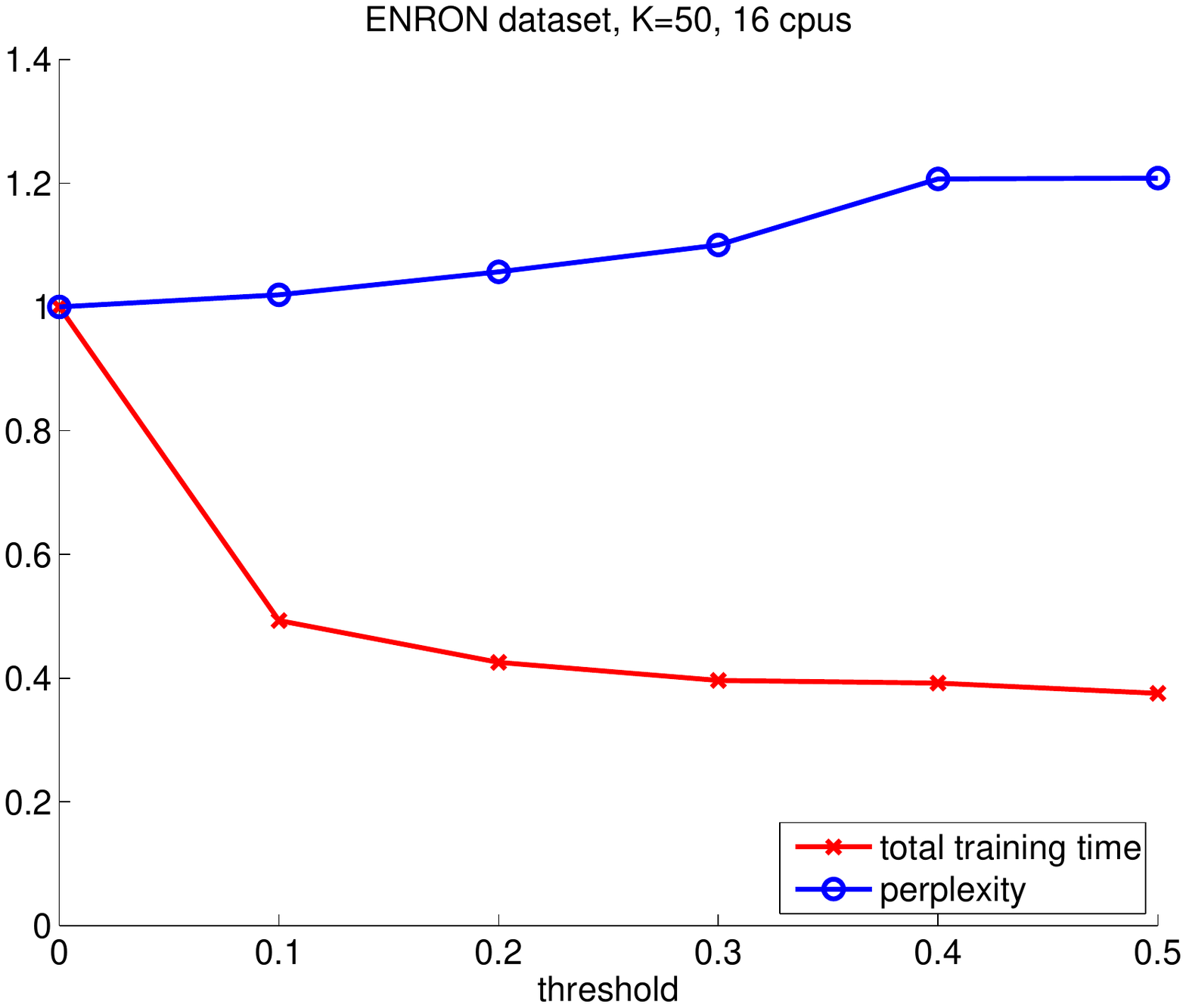}
}
\vspace{-30mm}
\centerline{
\includegraphics[width=0.45\textwidth]{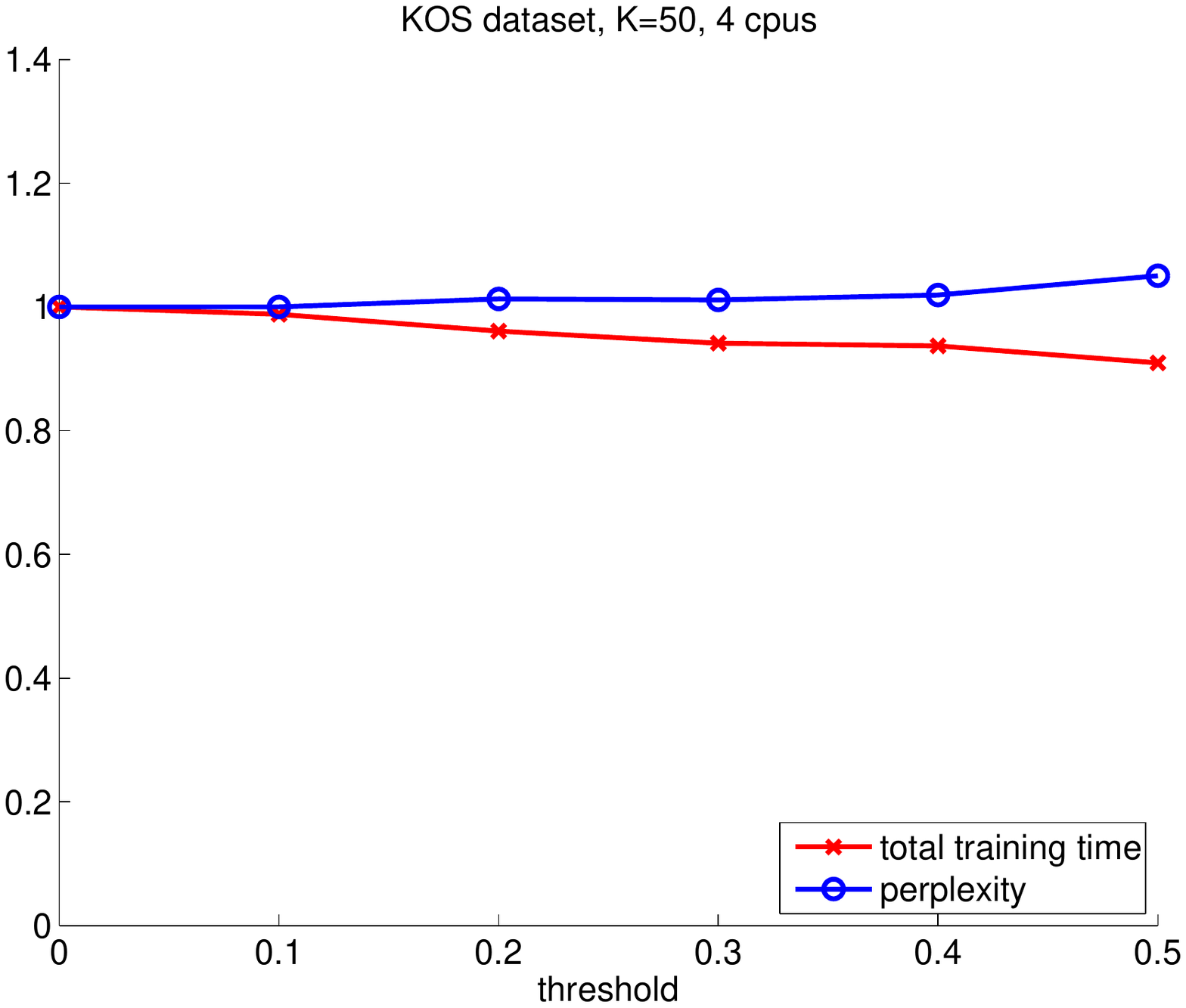}
\includegraphics[width=0.45\textwidth]{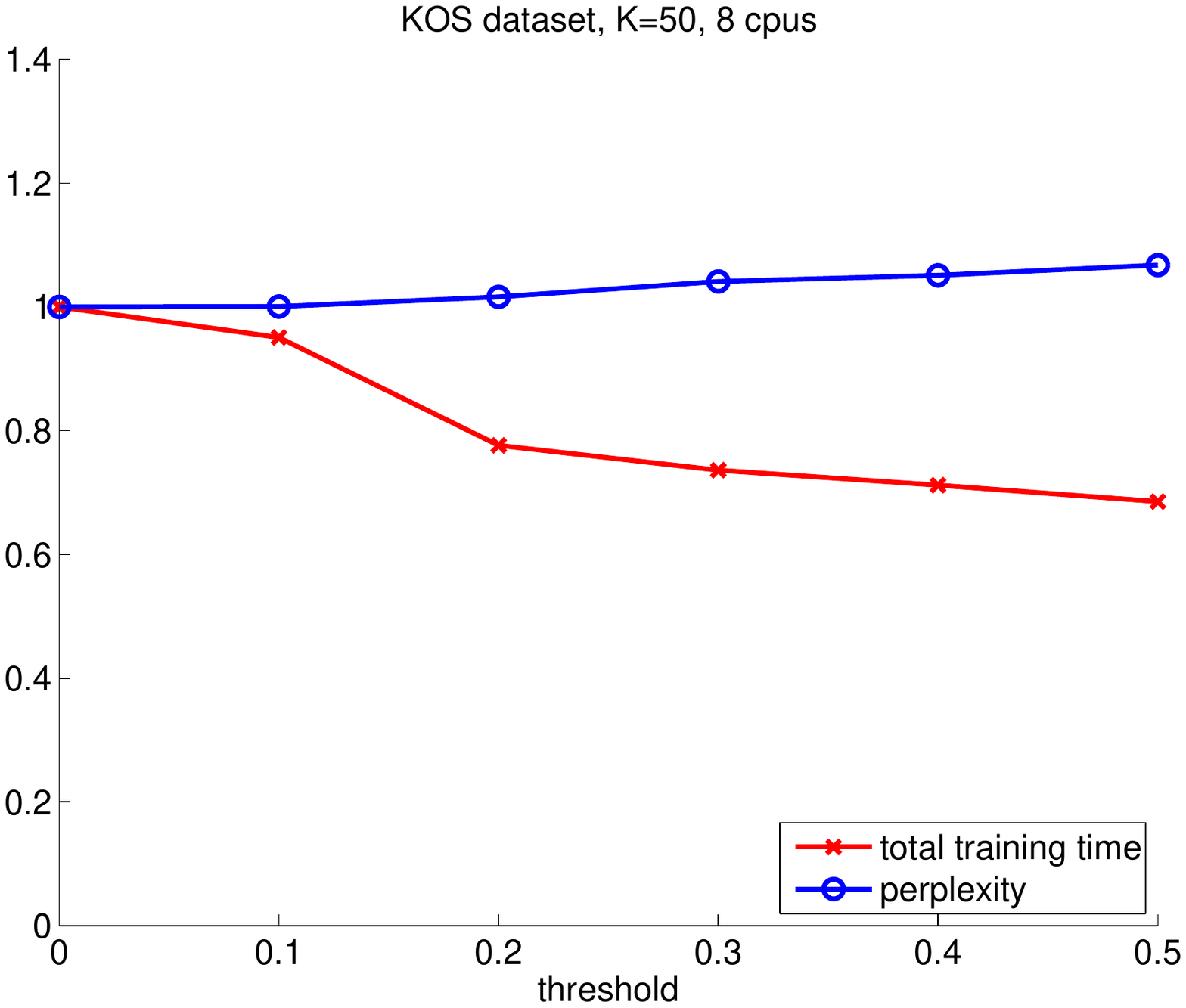}
\includegraphics[width=0.45\textwidth]{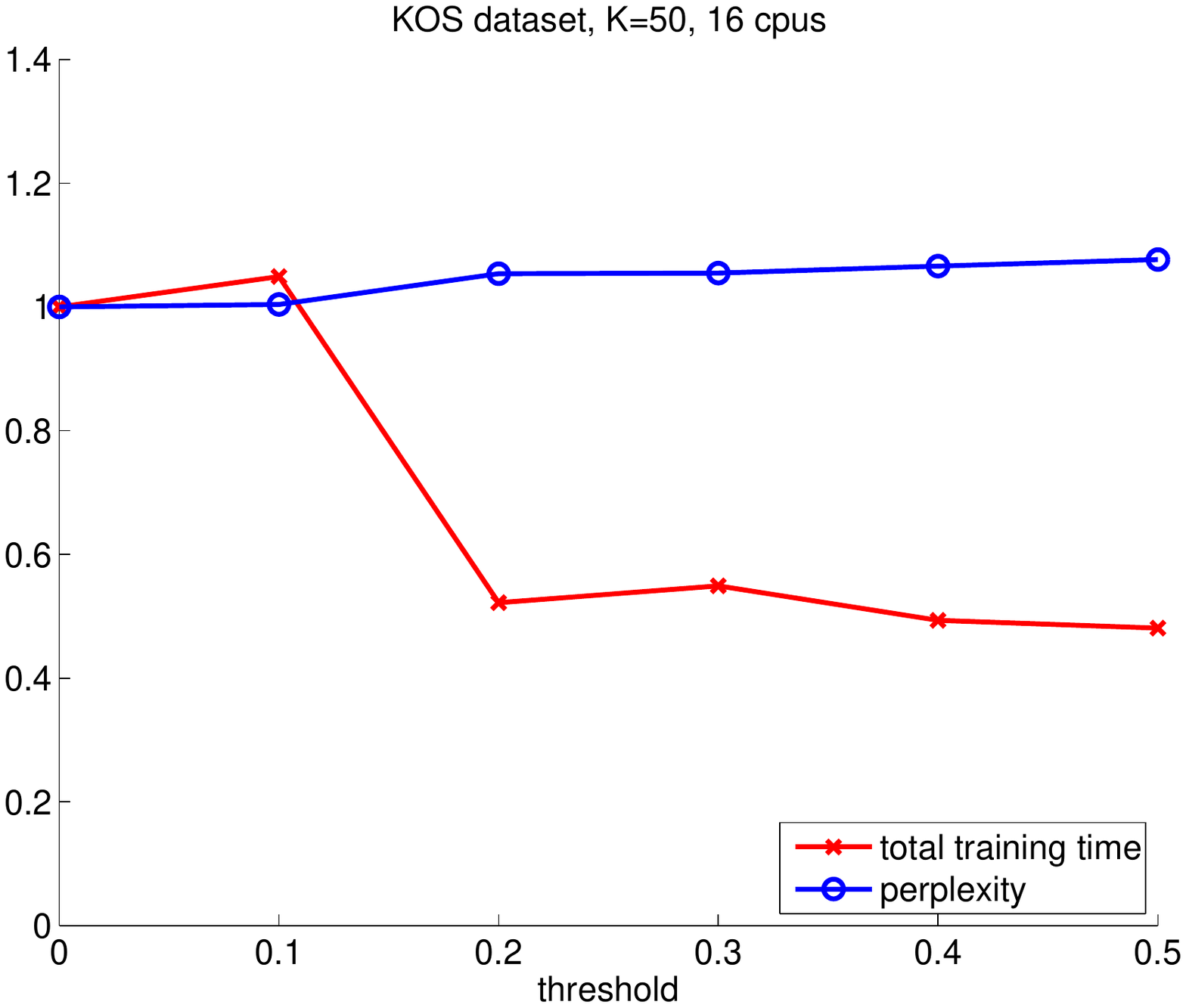}
}
\vspace{-15mm}
\caption{Normalized running time and test set perplexity as a function of the $threshold$ parameter. From left to right: 4, 8 and 16 CPUs. From top to bottom: NIPS, ENRON and KOS datasets.}
\label{fig:ttp_threshold}
\end{figure*}

In figure \ref{fig:ttp_threshold} we compare running time and perplexity for different values of $threshold$ and different number of CPUs. We can see that as we increase $threshold$ we can significantly reduce training time, with just a small impact on the quality of the approximation, measured by the \emph{perplexity} computed on a held-out test set. We can also see that, as expected, the training time reduction becomes more significant as we increasing the amount of information that has to be shared, by adding more CPUs to the task.

\begin{figure*}[tbp]
\vspace{-10mm}
\centerline{
\includegraphics[width=0.5\textwidth, height=0.4\textheight]{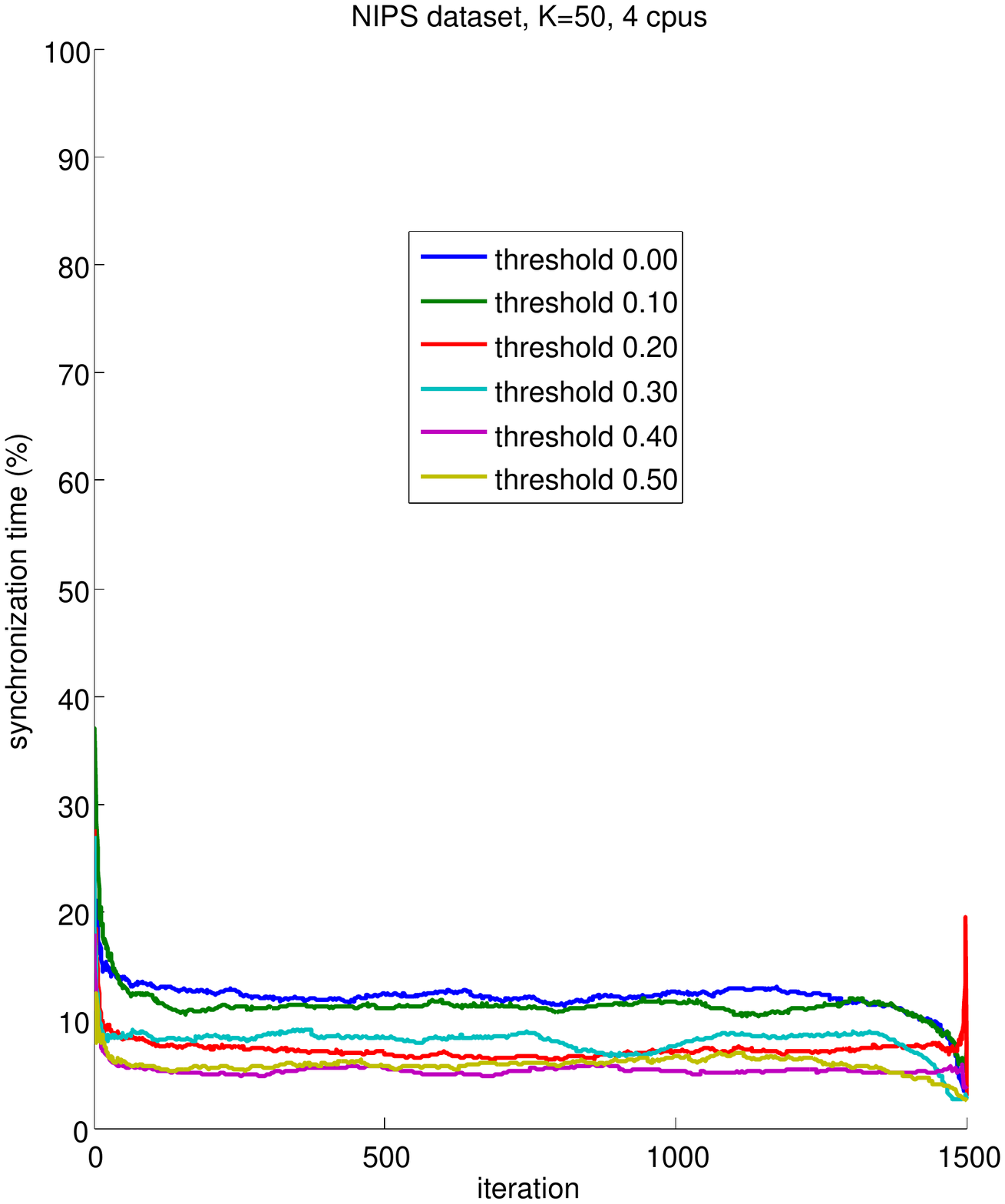}
\hspace{-13mm}
\includegraphics[width=0.5\textwidth, height=0.4\textheight]{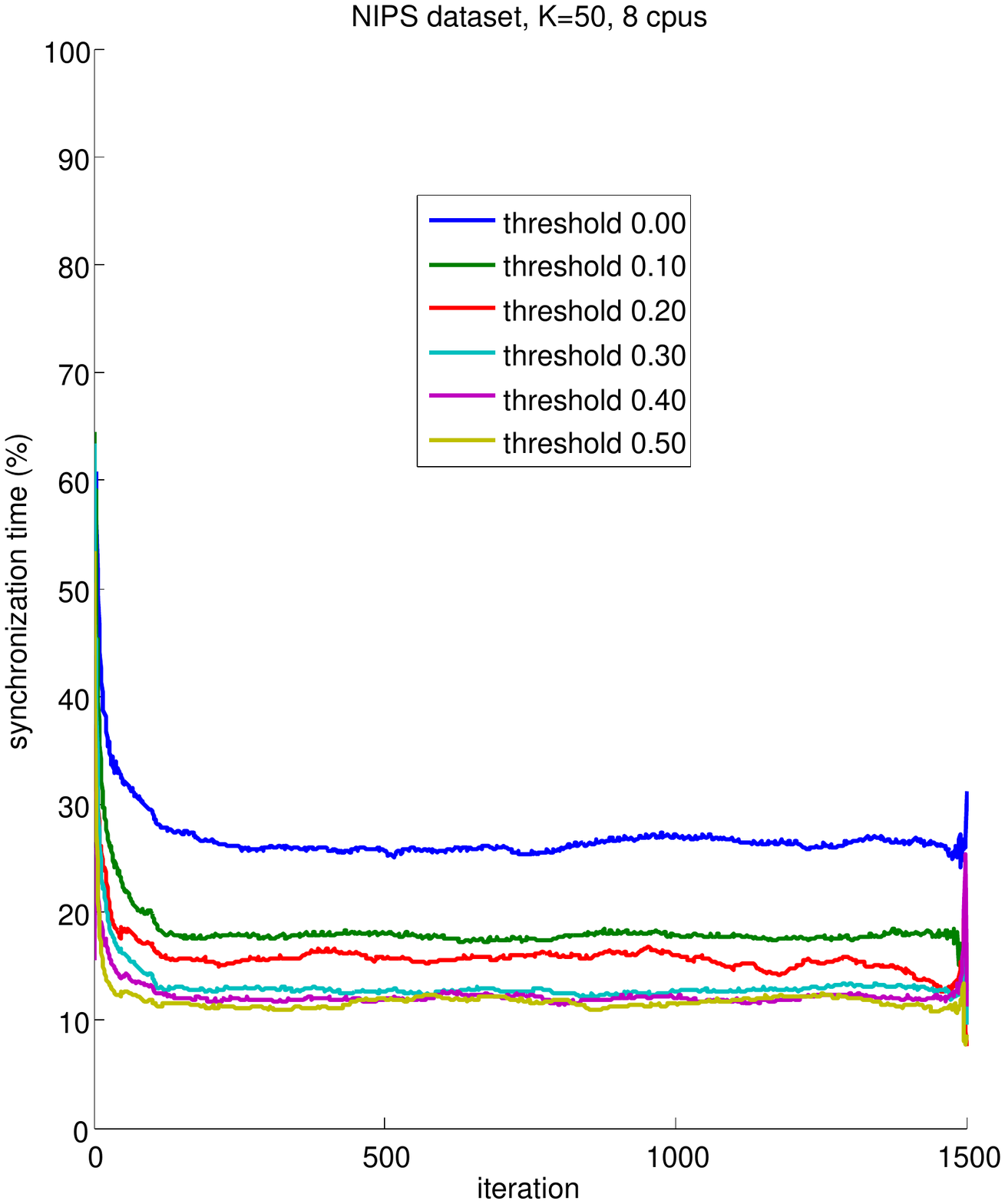}
\hspace{-13mm}
\includegraphics[width=0.5\textwidth, height=0.4\textheight]{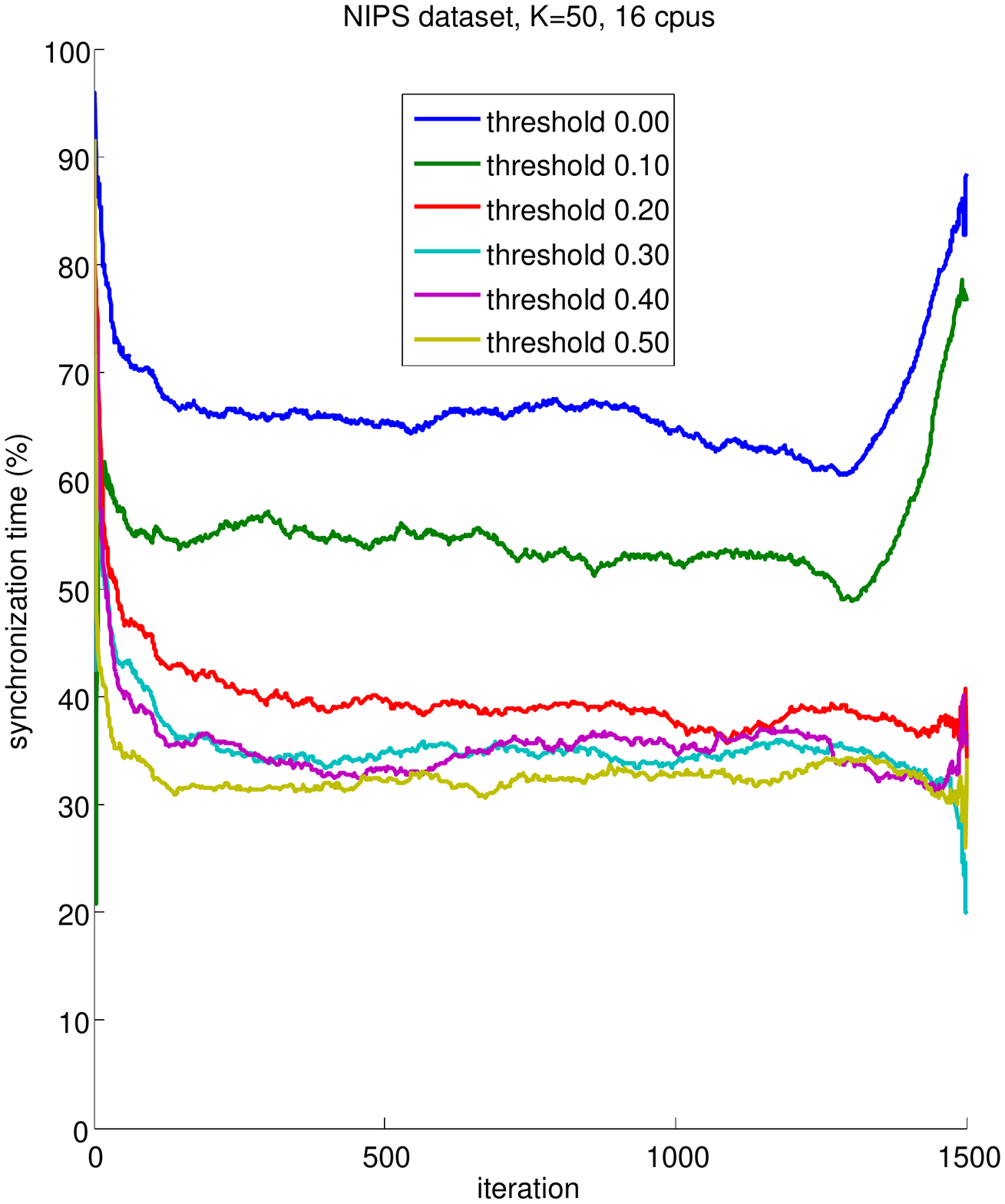}
}
\vspace{-20mm}
\centerline{
\includegraphics[width=0.5\textwidth, height=0.4\textheight]{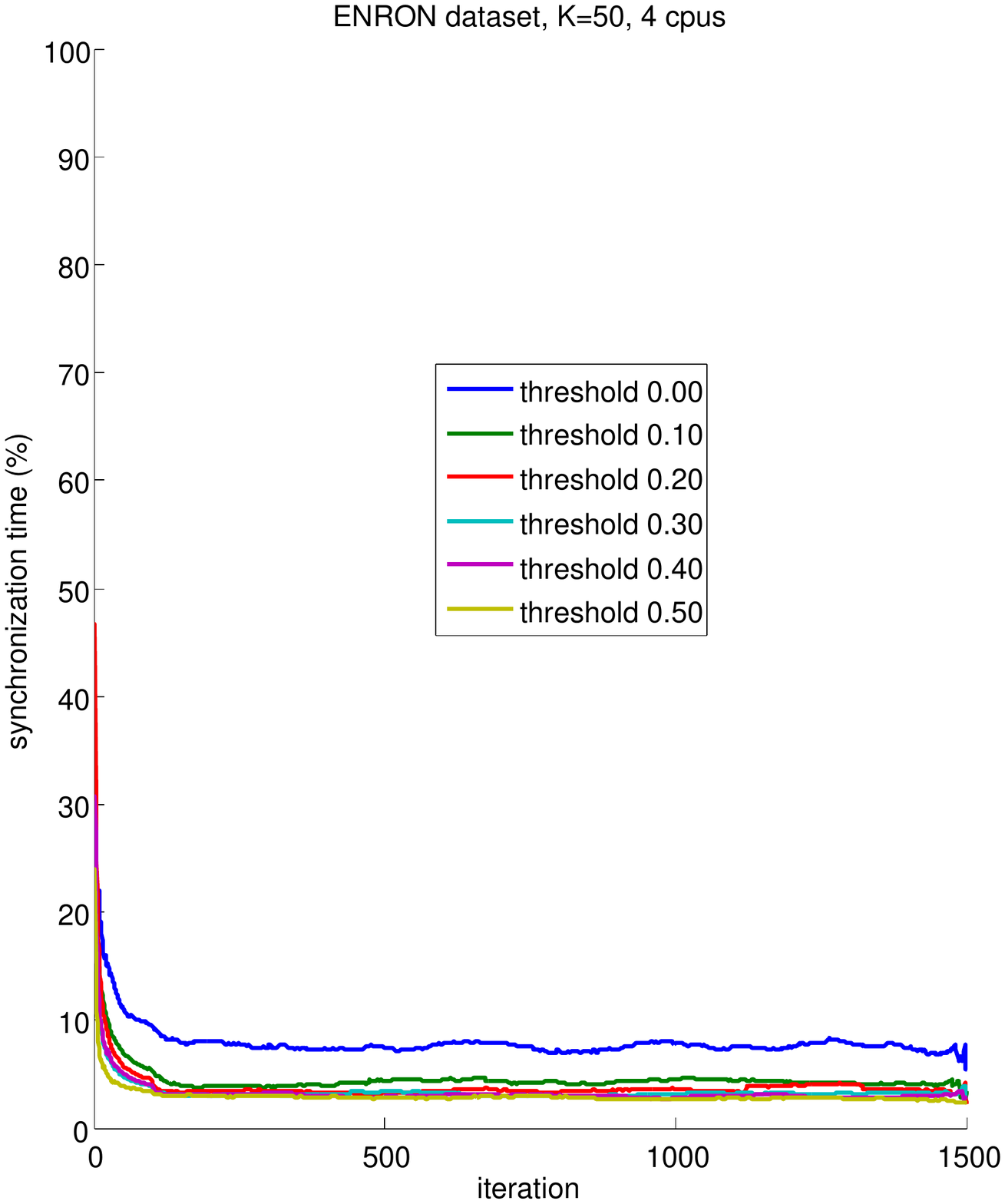}
\hspace{-13mm}
\includegraphics[width=0.5\textwidth, height=0.4\textheight]{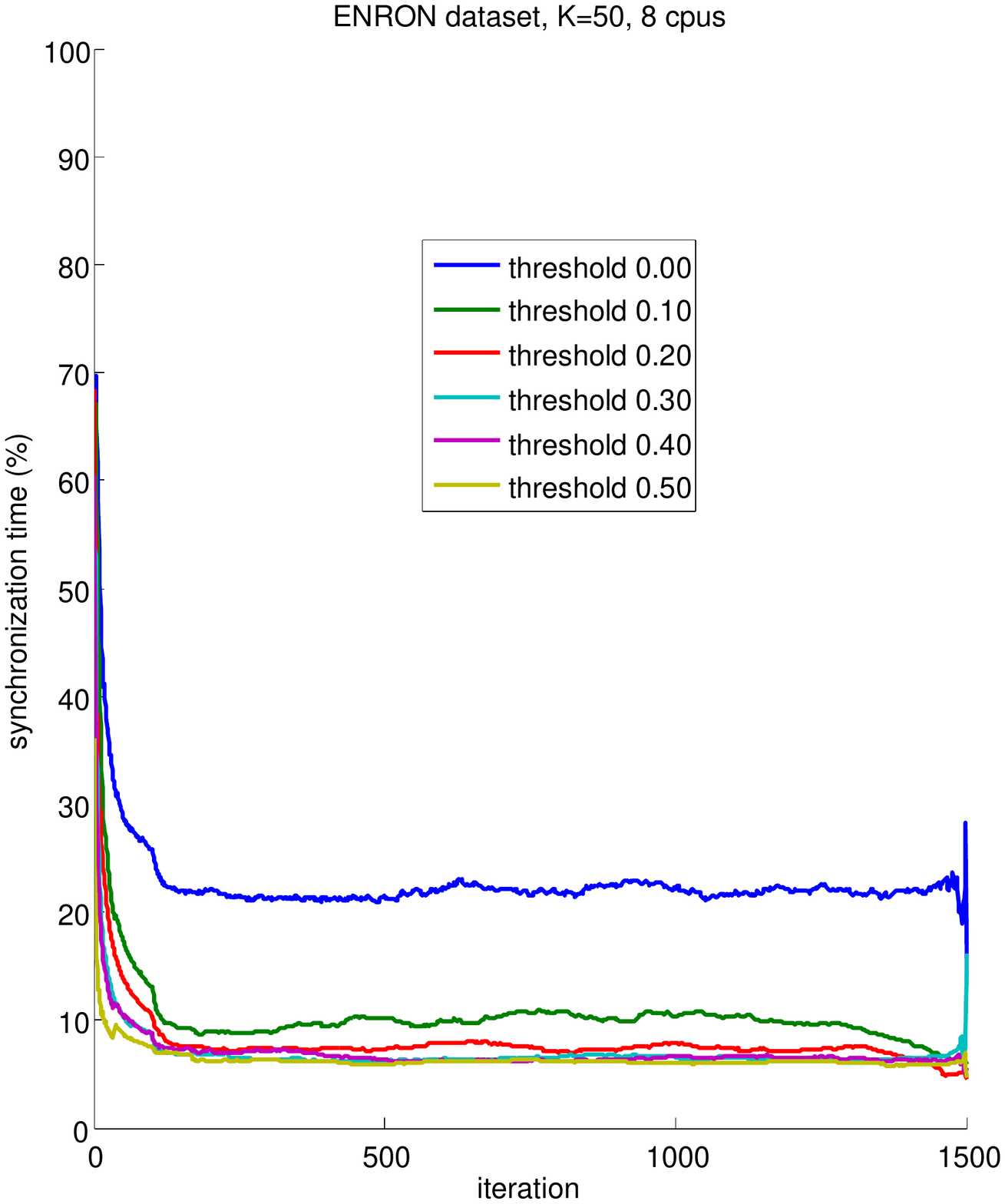}
\hspace{-13mm}
\includegraphics[width=0.5\textwidth, height=0.4\textheight]{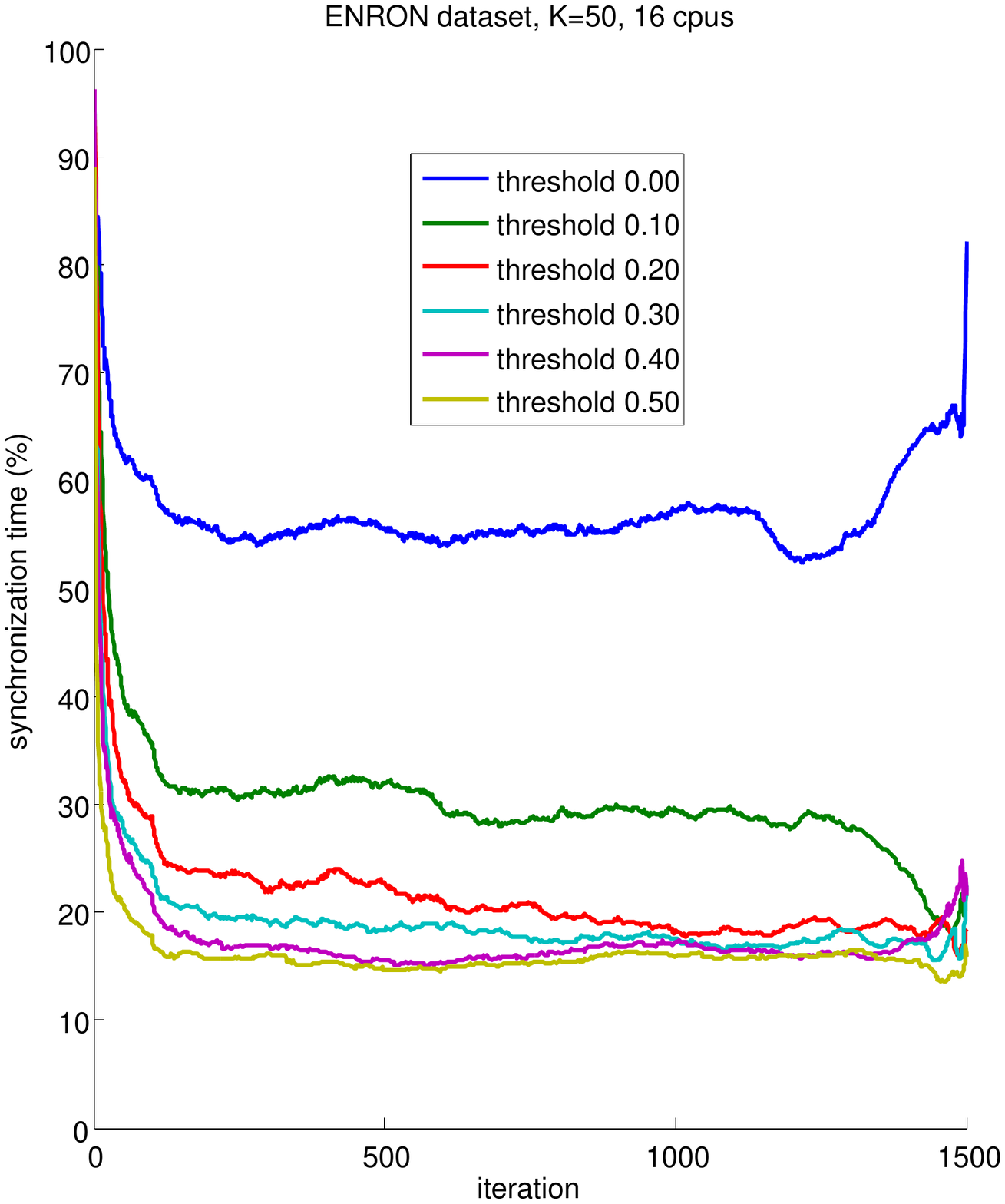}
}
\vspace{-20mm}
\centerline{
\includegraphics[width=0.5\textwidth, height=0.4\textheight]{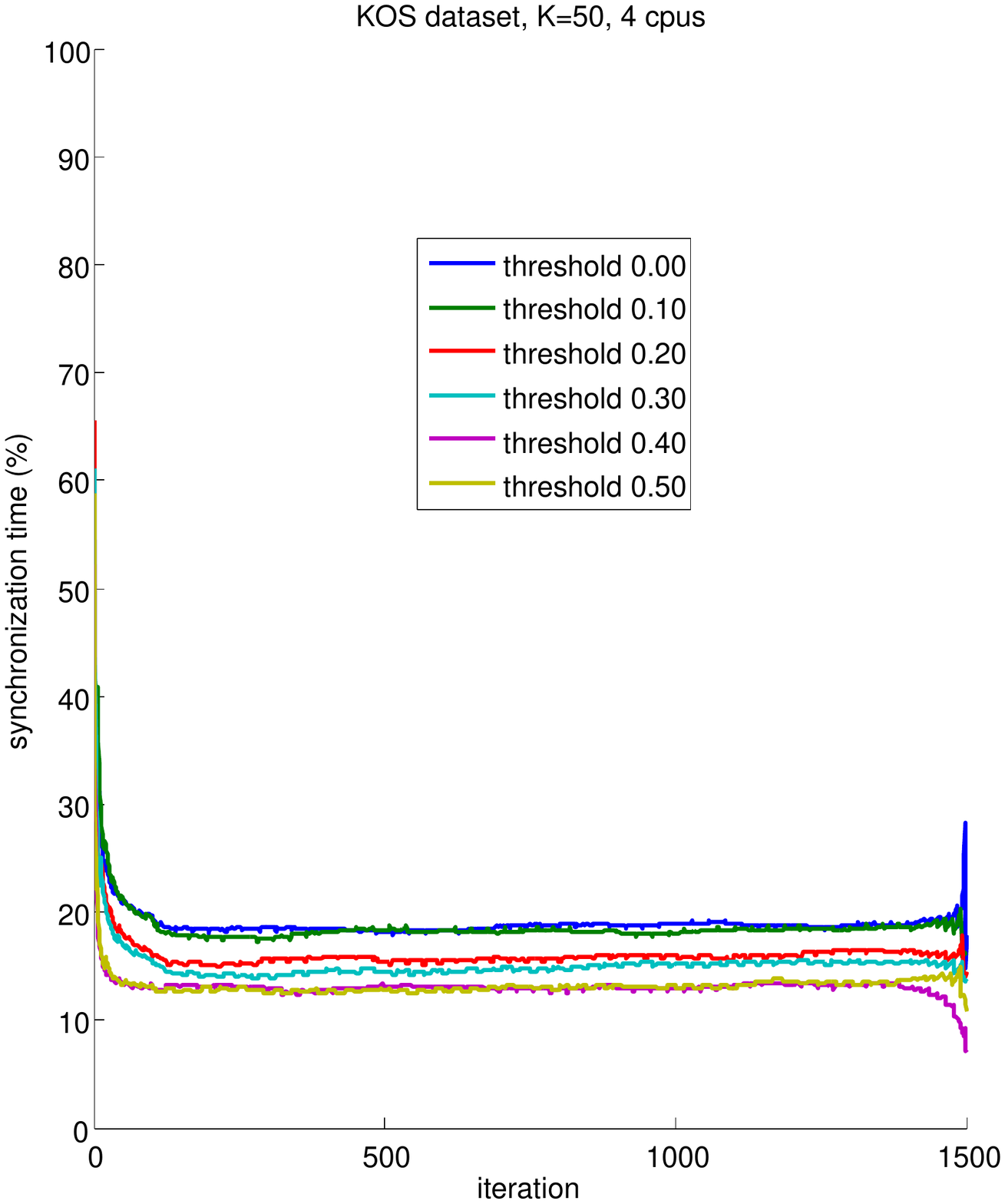}
\hspace{-13mm}
\includegraphics[width=0.5\textwidth, height=0.4\textheight]{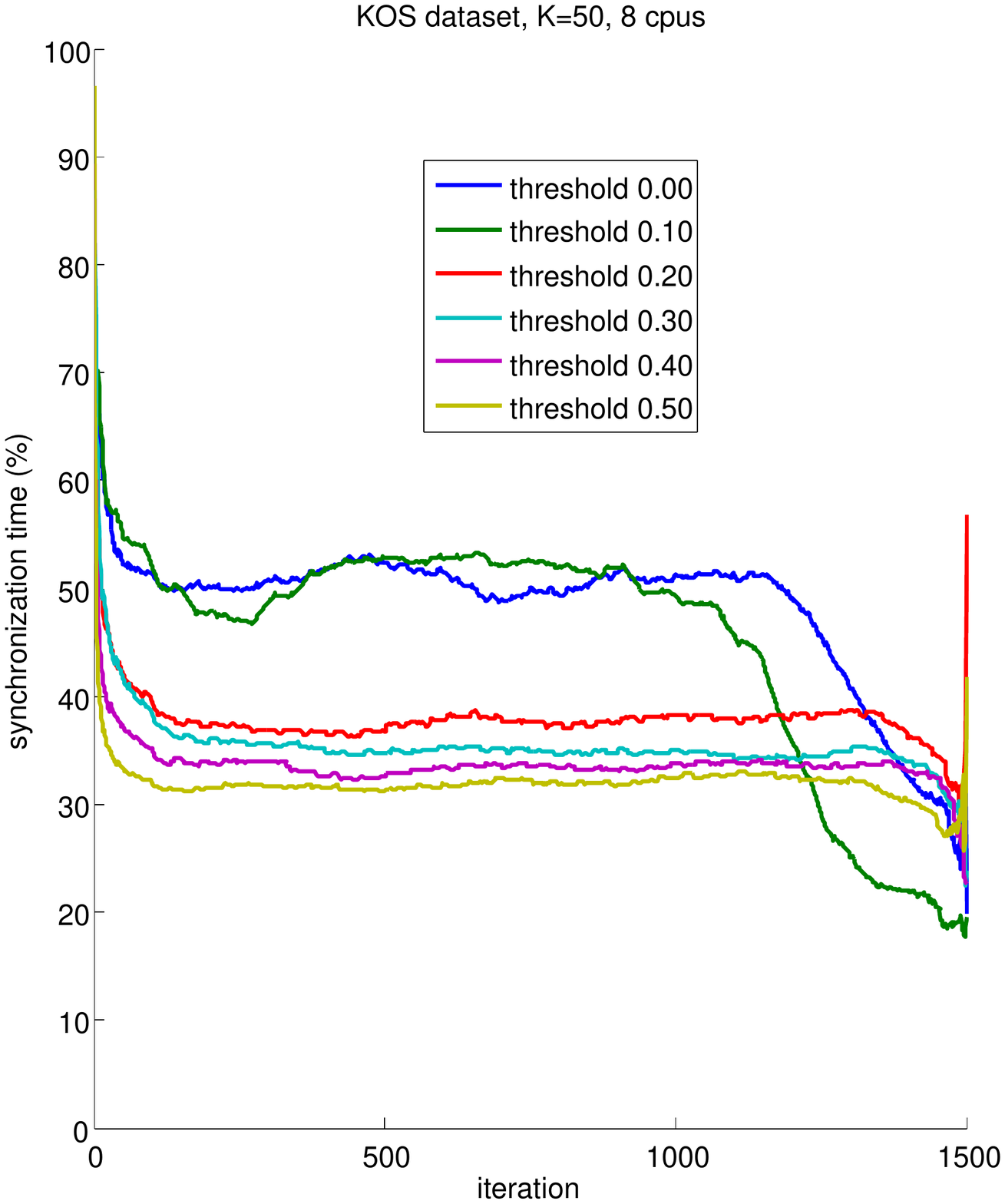}
\hspace{-13mm}
\includegraphics[width=0.5\textwidth, height=0.4\textheight]{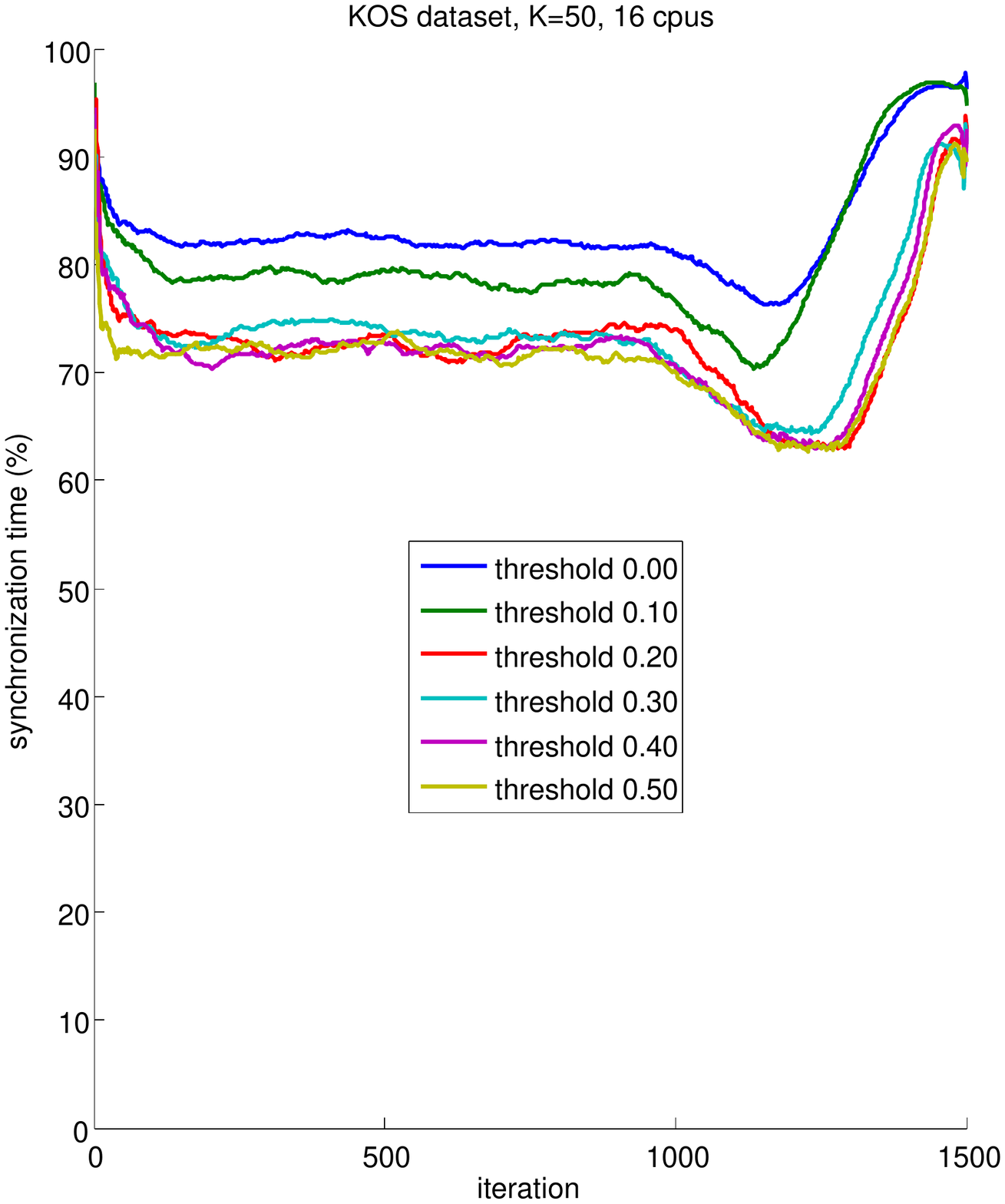}
}
\vspace{-7mm}
\caption{Proportion of iteration time spent with synchronization for different values of $threshold$ (this plot was smoothed with a moving average filter with a span of 200 iterations). From left to right: 4, 8 and 16 CPUs. From top to bottom: NIPS, ENRON and KOS datasets.}
\label{fig:sync_threshold}
\end{figure*}

In figure \ref{fig:sync_threshold} we show the proportion of time spent in synchronization at each iteration when training the LDA model with different numbers of CPUs. By increasing $threshold$ we can substantially decrease synchronization time. As expected, as the number of CPUs increase synchronization starts to dominate over processing time.

\begin{figure*}[tbp]
\centerline{
\includegraphics[width=0.5\textwidth]{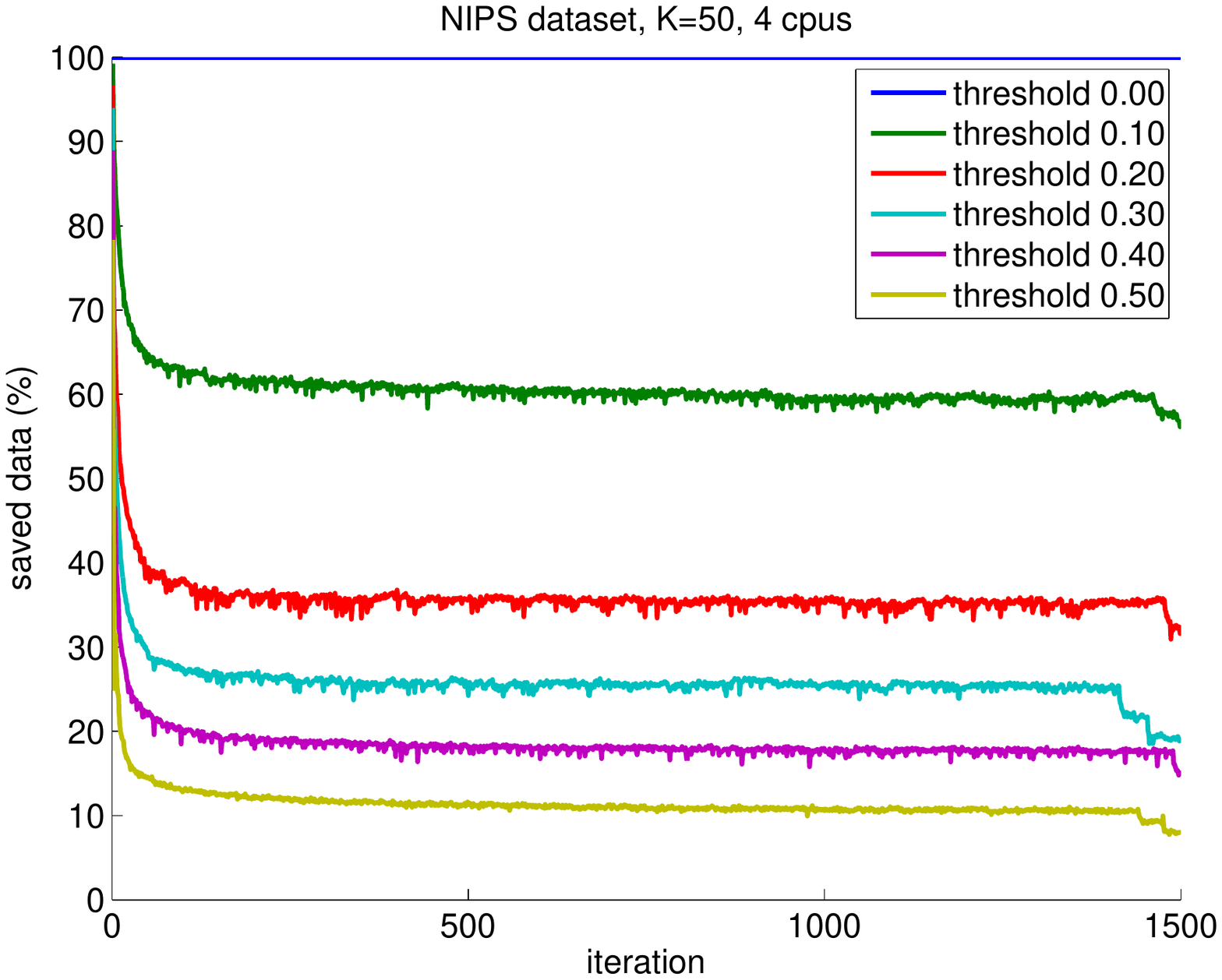}
\hspace{-13mm}
\includegraphics[width=0.5\textwidth]{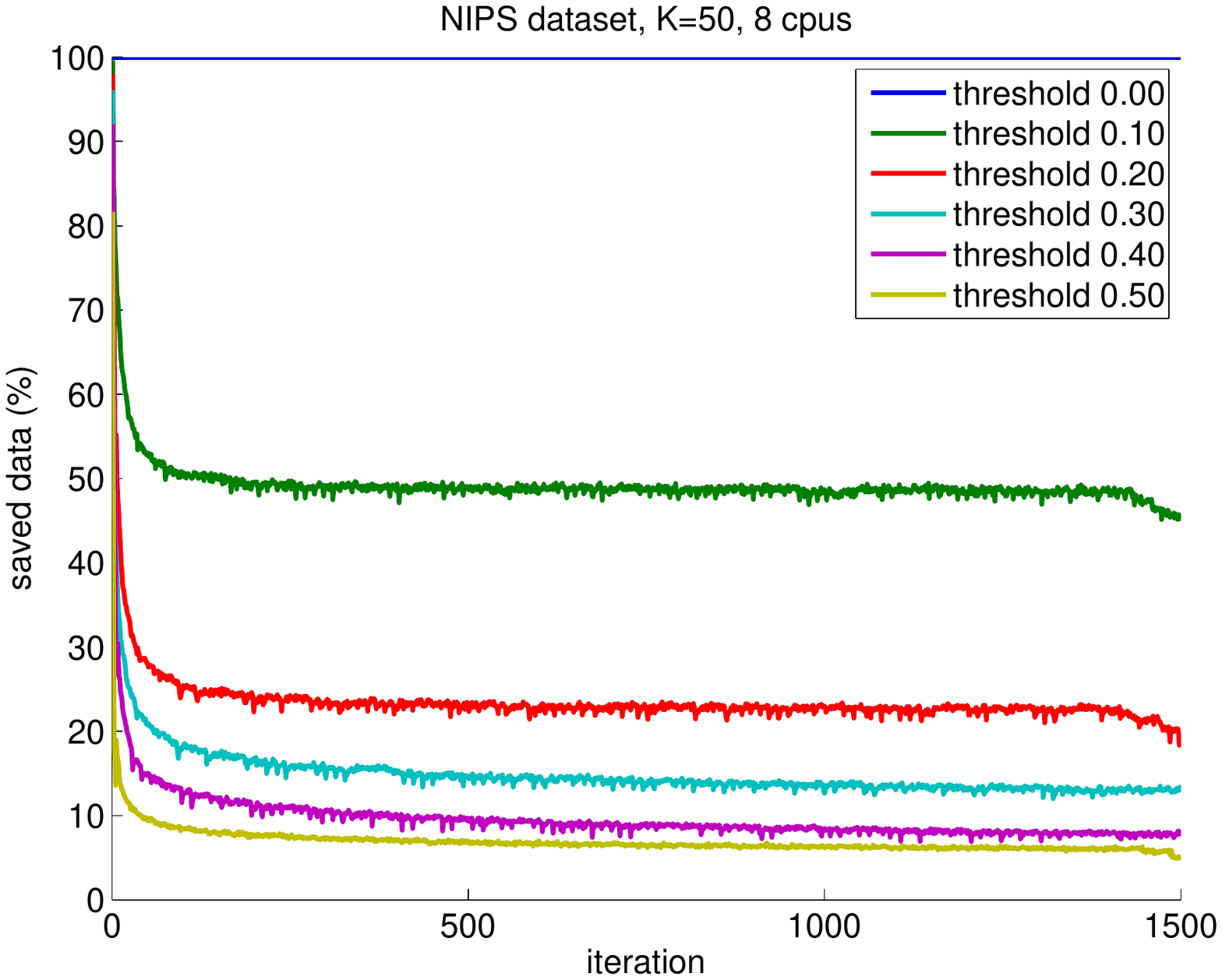}
\hspace{-13mm}
\includegraphics[width=0.5\textwidth]{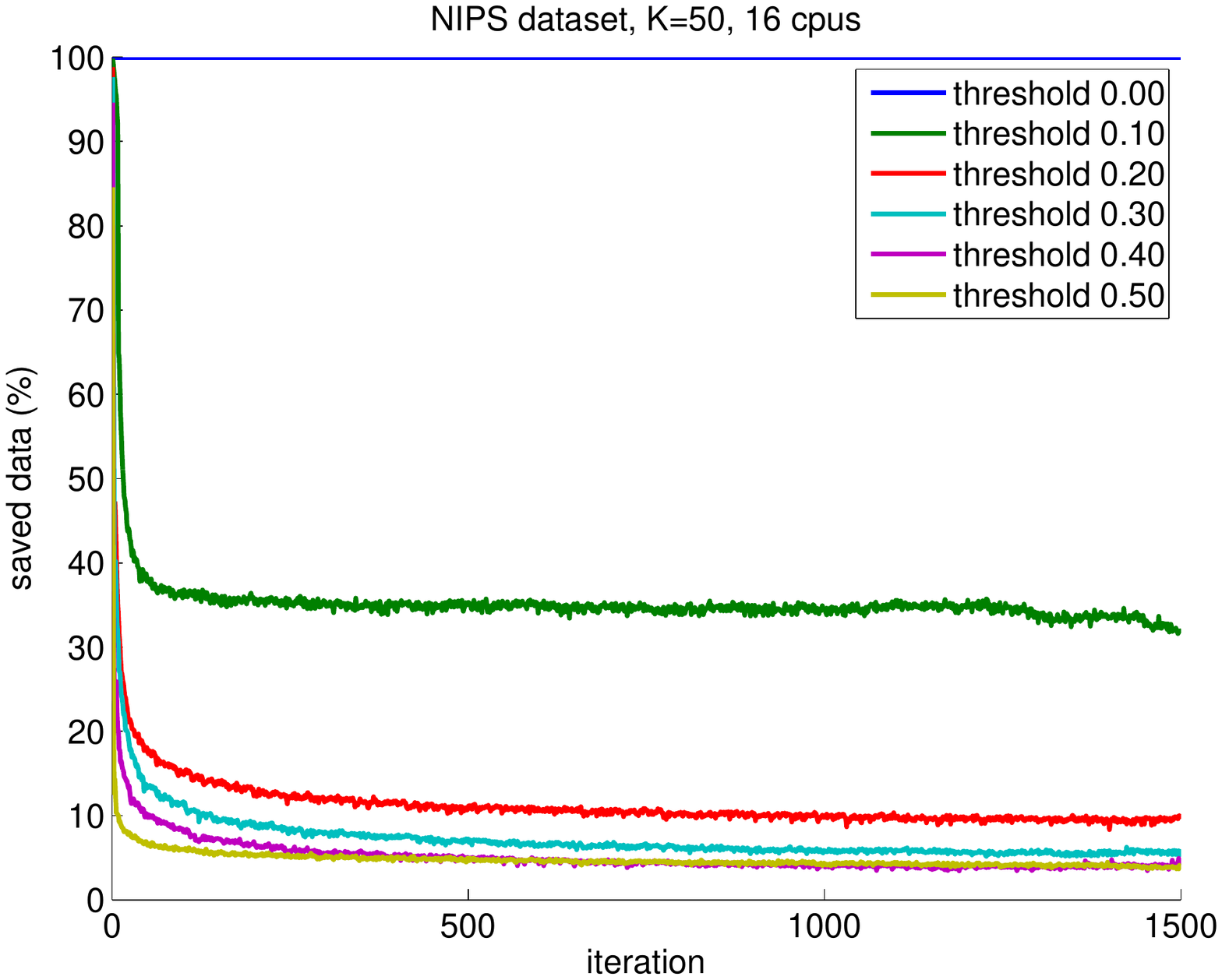}
}
\vspace{-40mm}
\centerline{
\includegraphics[width=0.5\textwidth]{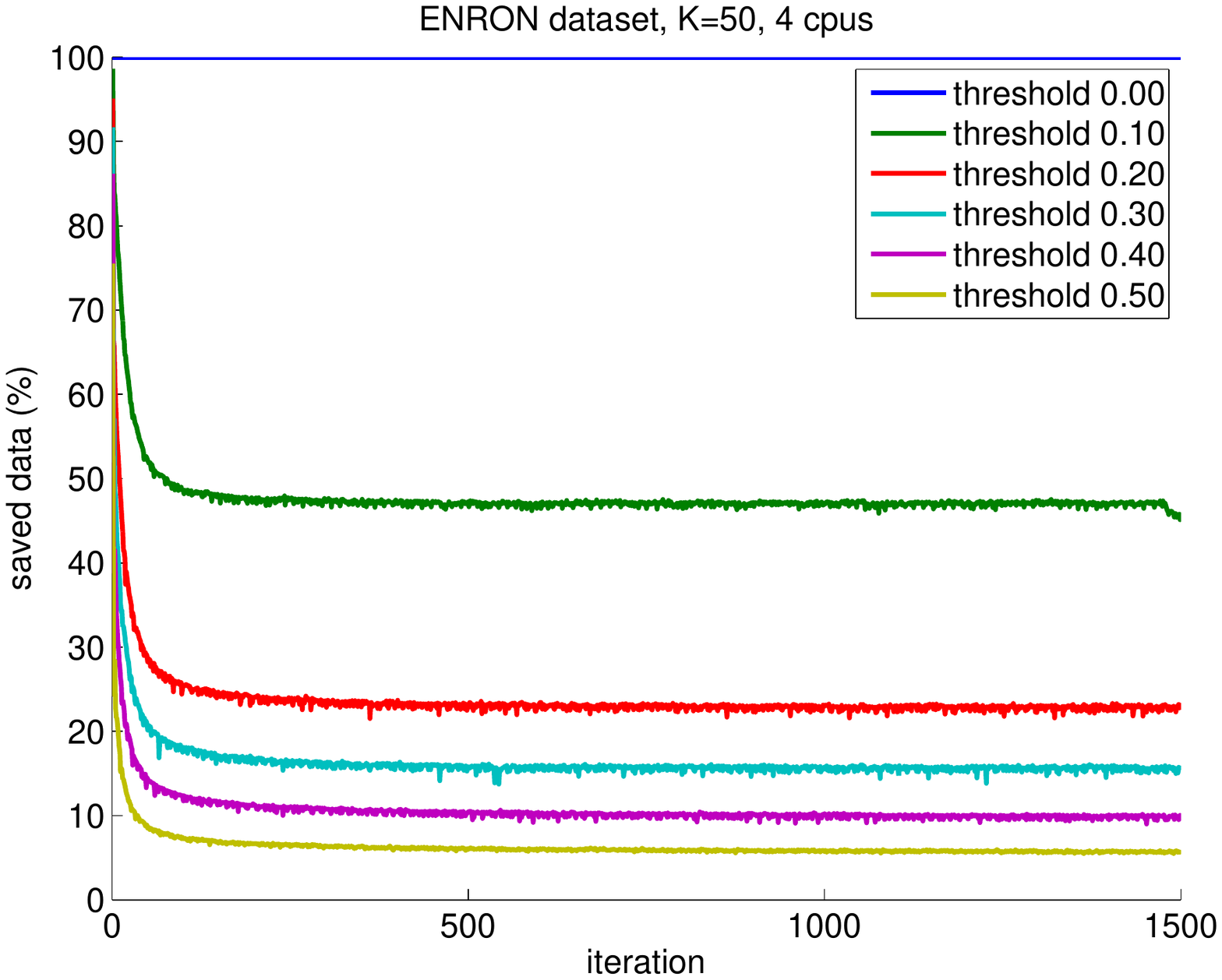}
\hspace{-13mm}
\includegraphics[width=0.5\textwidth]{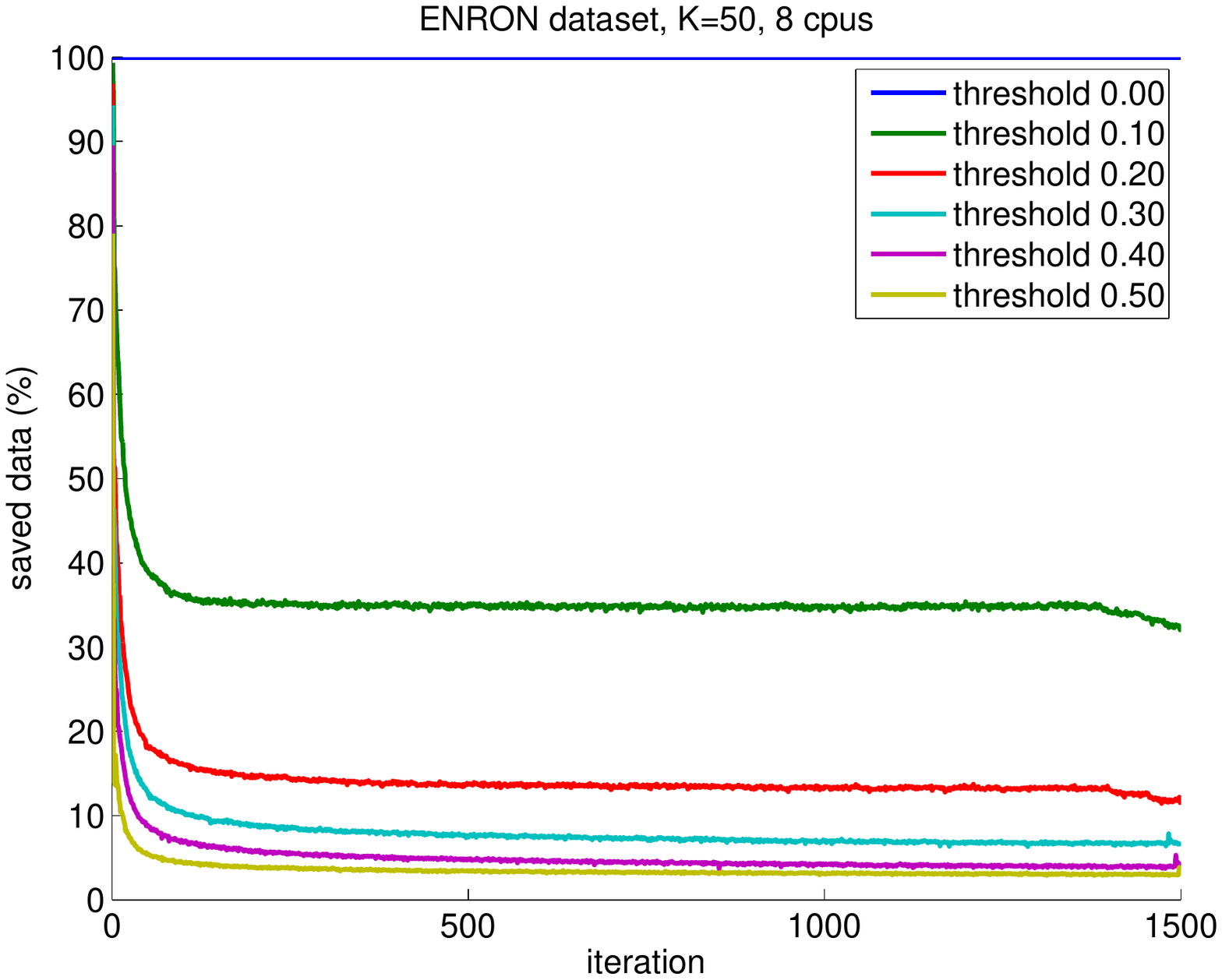}
\hspace{-13mm}
\includegraphics[width=0.5\textwidth]{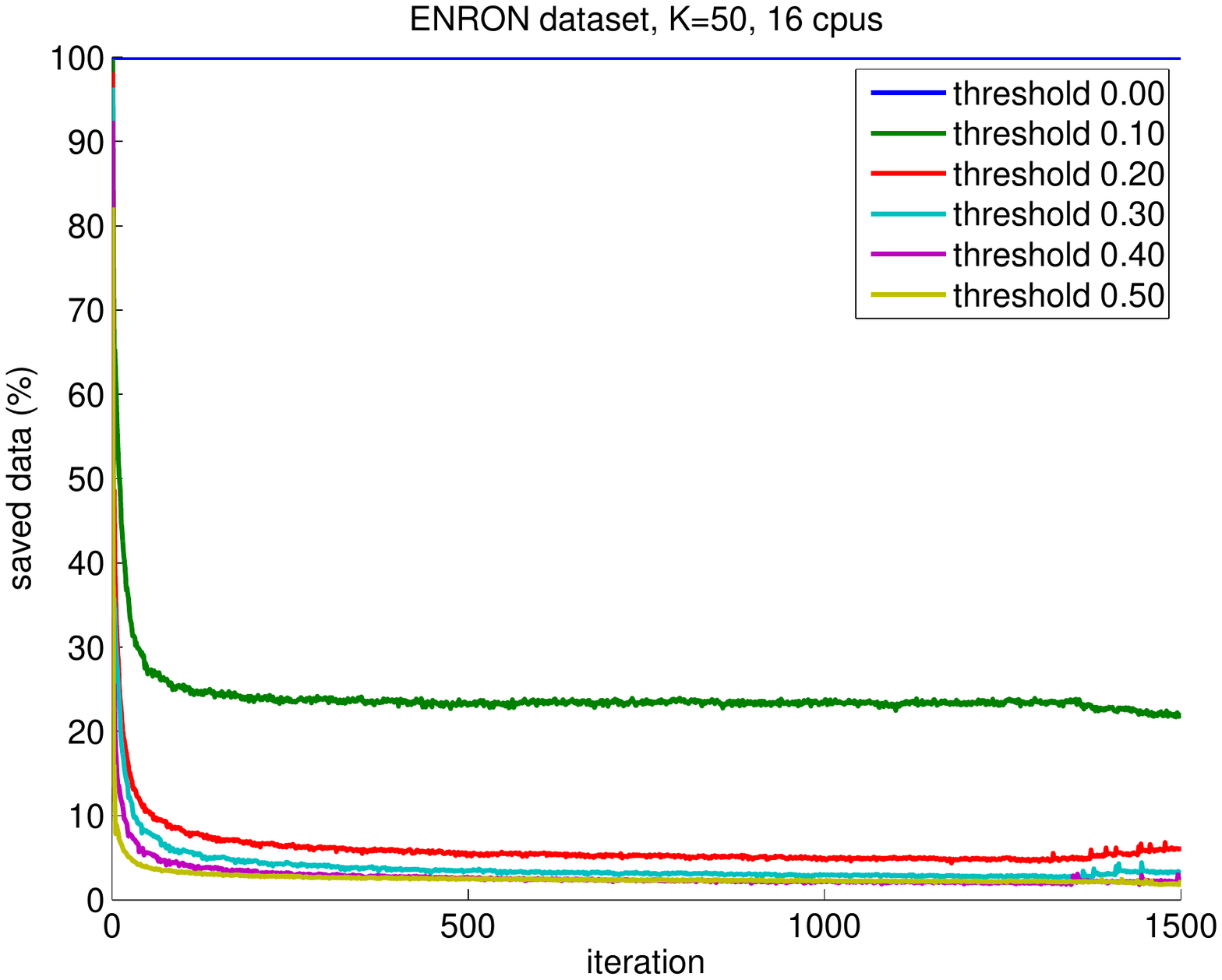}
}
\vspace{-40mm}
\centerline{
\includegraphics[width=0.5\textwidth]{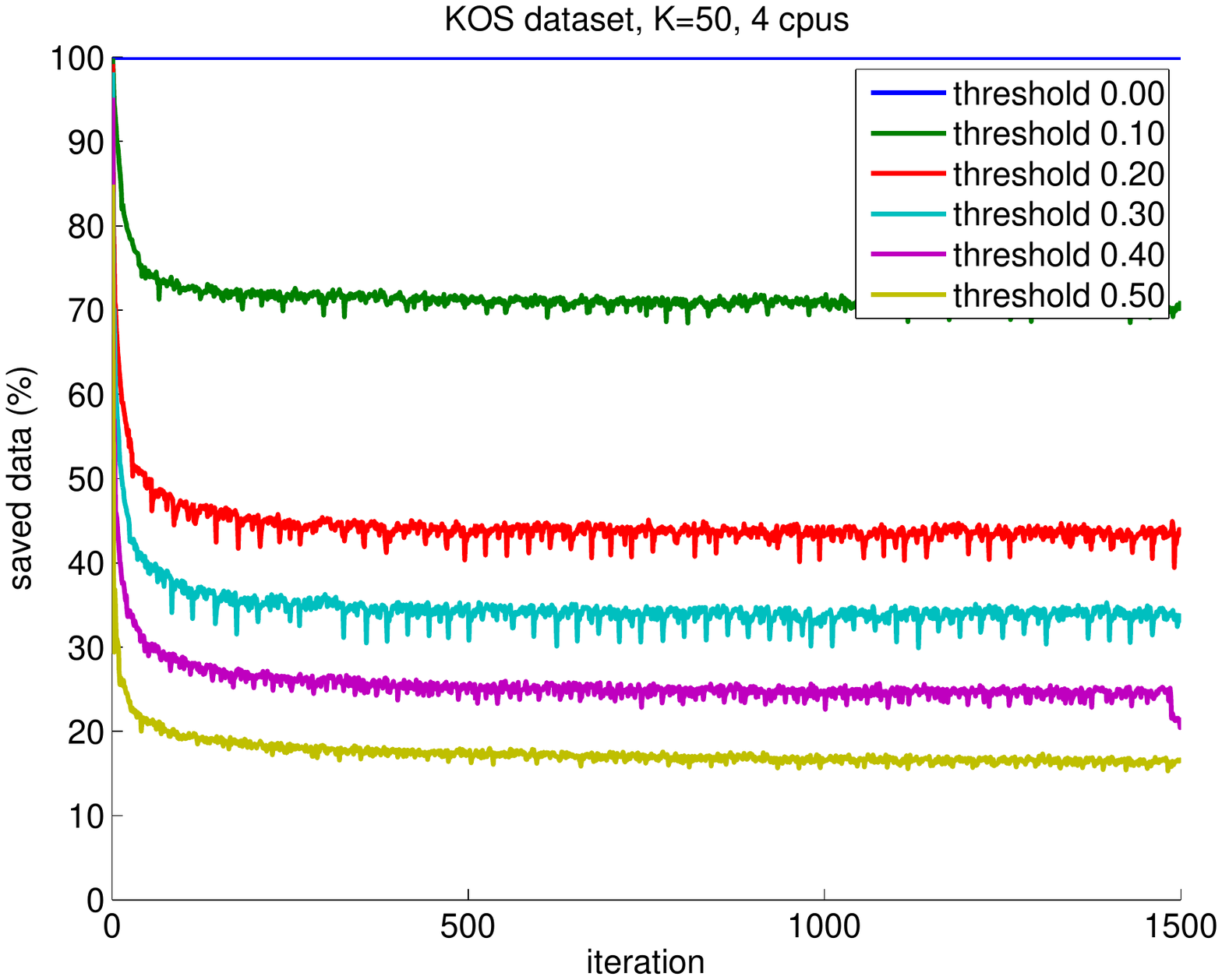}
\hspace{-13mm}
\includegraphics[width=0.5\textwidth]{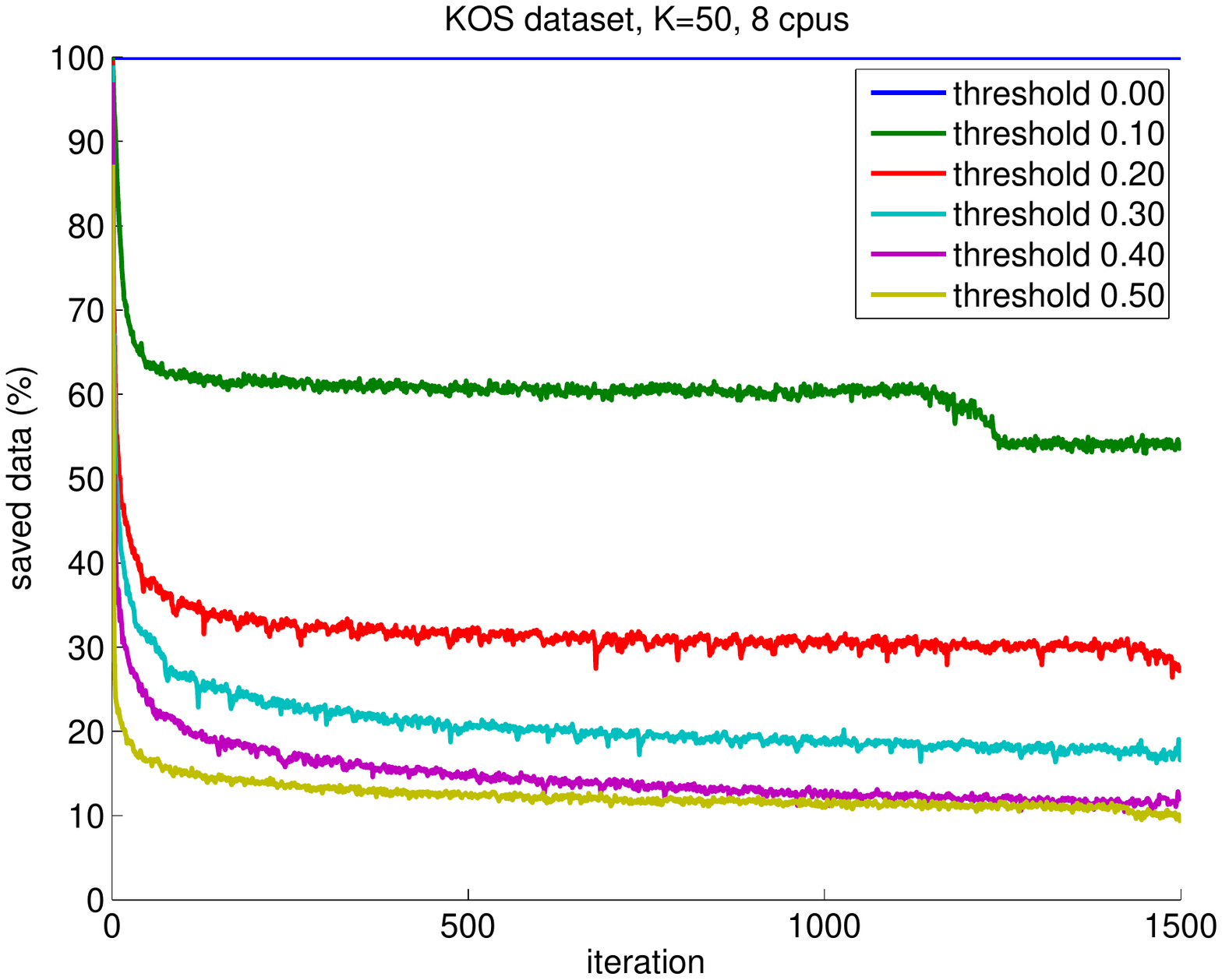}
\hspace{-13mm}
\includegraphics[width=0.5\textwidth]{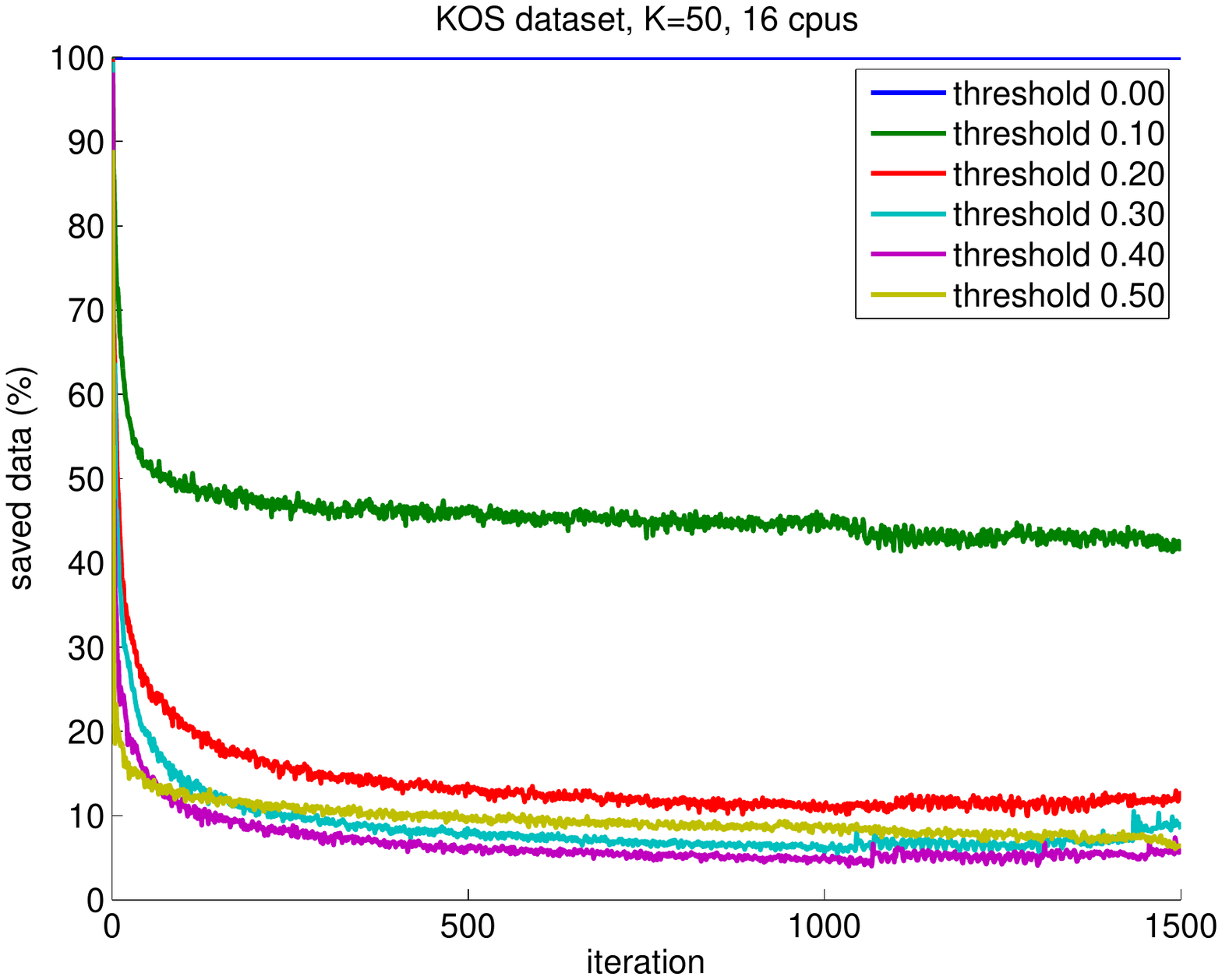}
}
\vspace{-10mm}
\caption{Proportion of local modifications to $n_{k,v}$ saved at each iteration, for different values of $threshold$. From left to right: 4, 8 and 16 CPUs. From top to bottom: NIPS, ENRON and KOS datasets.}
\label{fig:saved_threshold}
\end{figure*}

In figure \ref{fig:saved_threshold} we show the amount of information saved at each step for different values of $threshold$. We see that in the first few iterations the savings obtained by Algorithm \ref{alg:sparse} are small, since almost all modifications are \emph{relevant}, but as the model converges the amount of \emph{relevant} information stabilizes at a lower level. We can also see that as we add more CPUs the savings become more prominent -- this is expected, since then modifications of a single CPU tend to be less relevant as it becomes responsible for a smaller proportion of the corpus.

\begin{figure*}[tbp]
\centerline{
\includegraphics[width=0.5\textwidth]{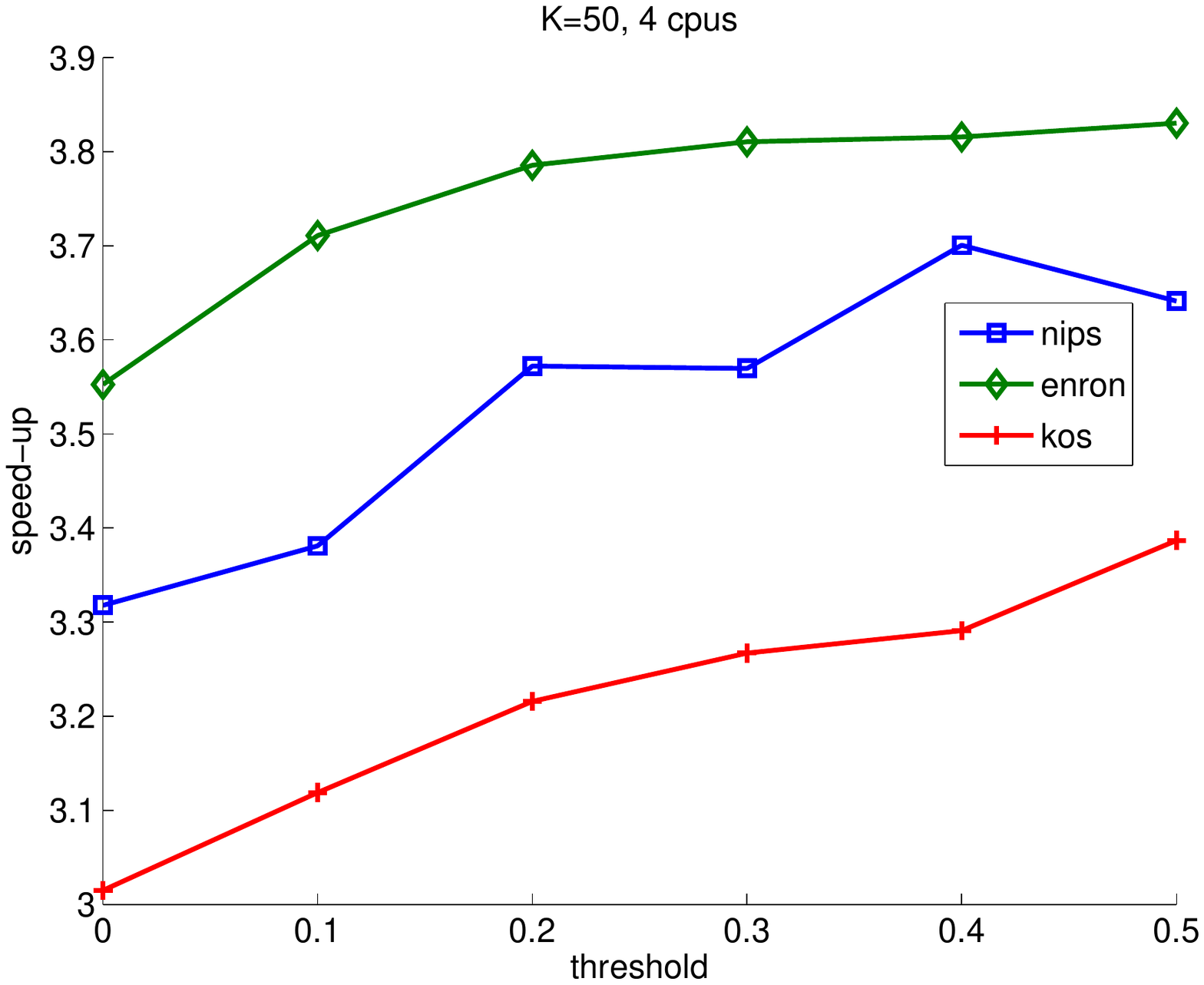}
\hspace{-10mm}
\includegraphics[width=0.5\textwidth]{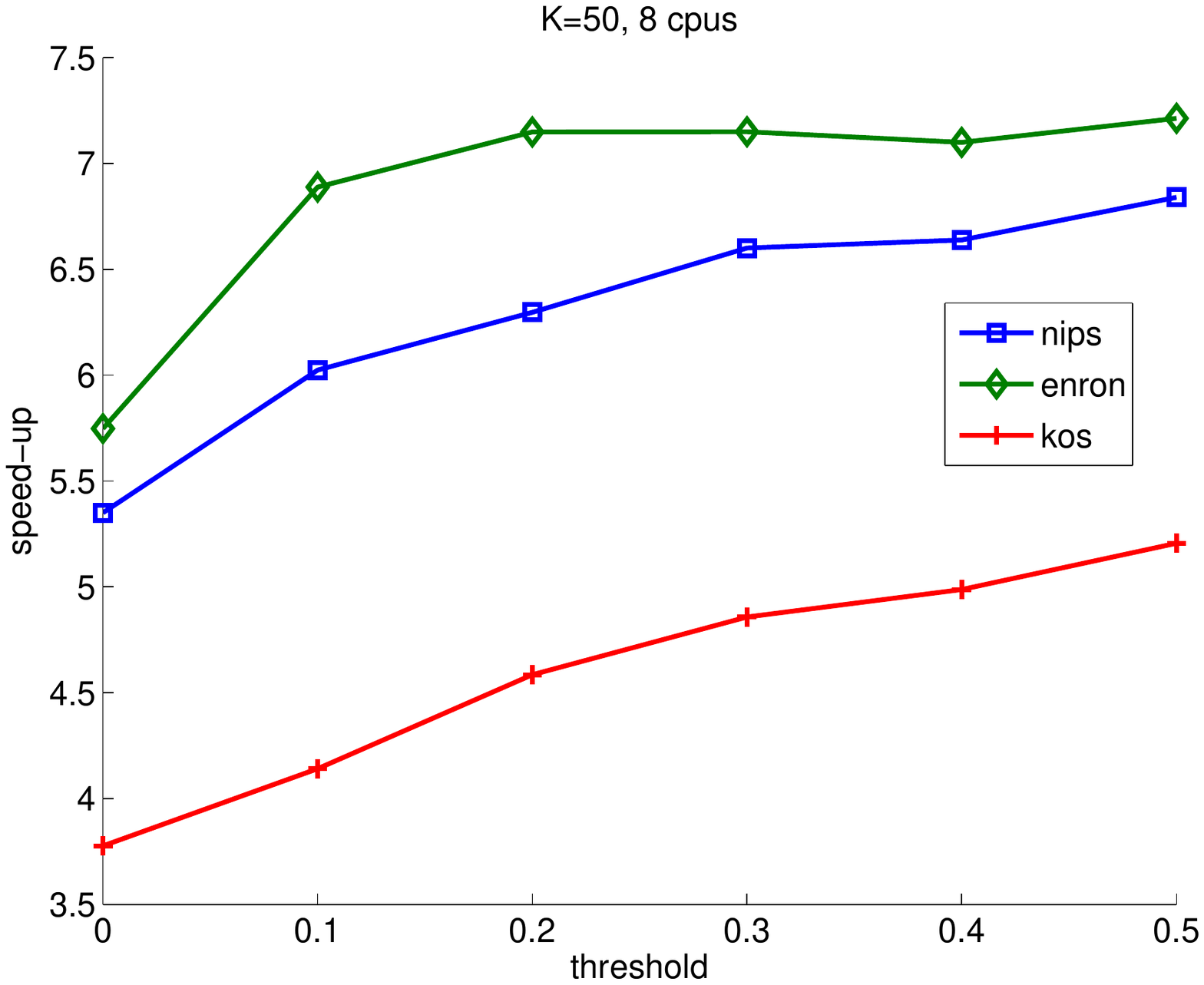}
\hspace{-10mm}
\includegraphics[width=0.5\textwidth]{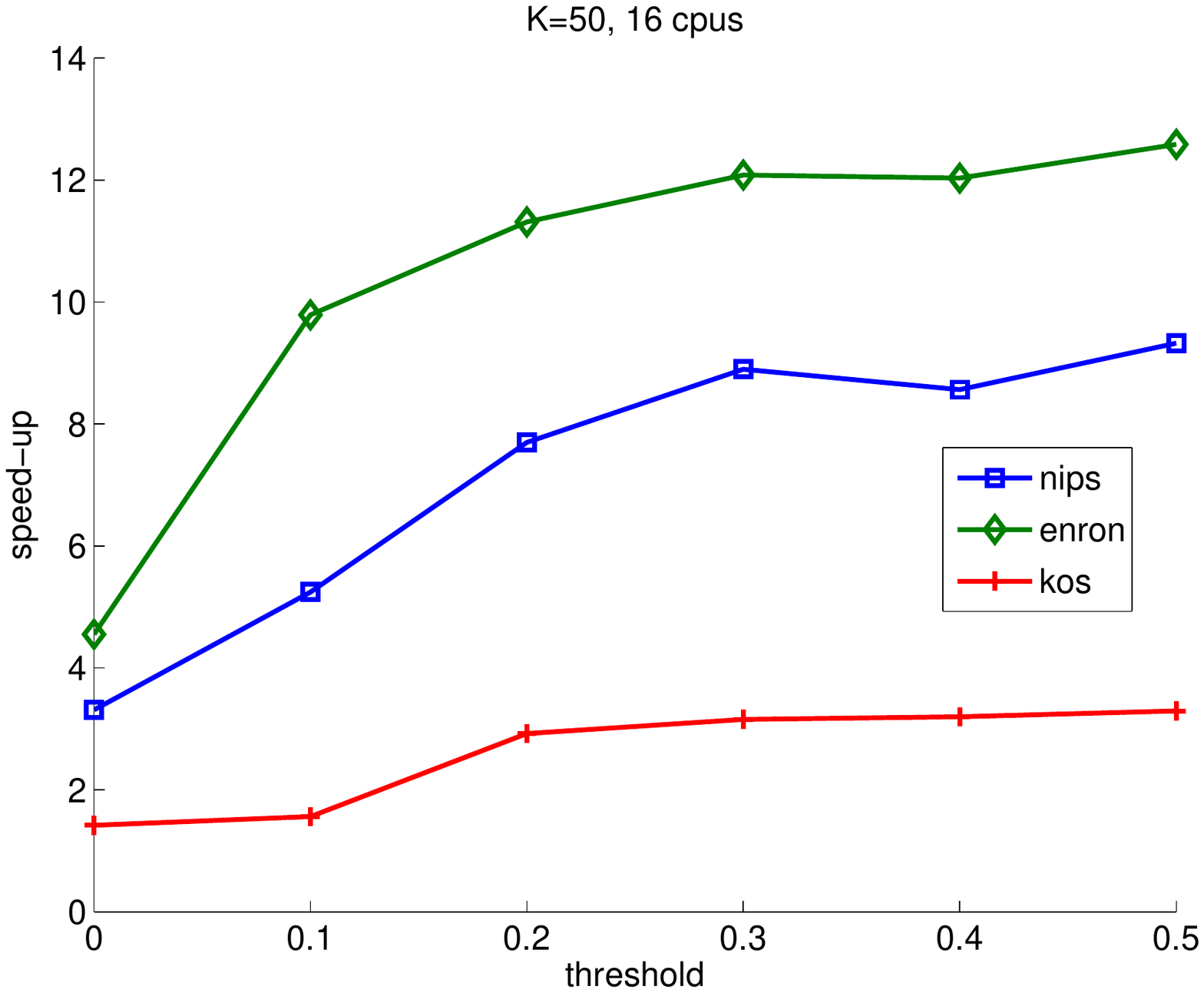}
}
\vspace{-20mm}
\caption{Speed-up compared to one-core implementation for different values of $threshold$. From left to right: 4, 8 and 16 CPUs.}
\label{fig:speed_threshold}
\end{figure*}

\begin{figure*}[tbp]
\centerline{
\includegraphics[width=0.6\textwidth]{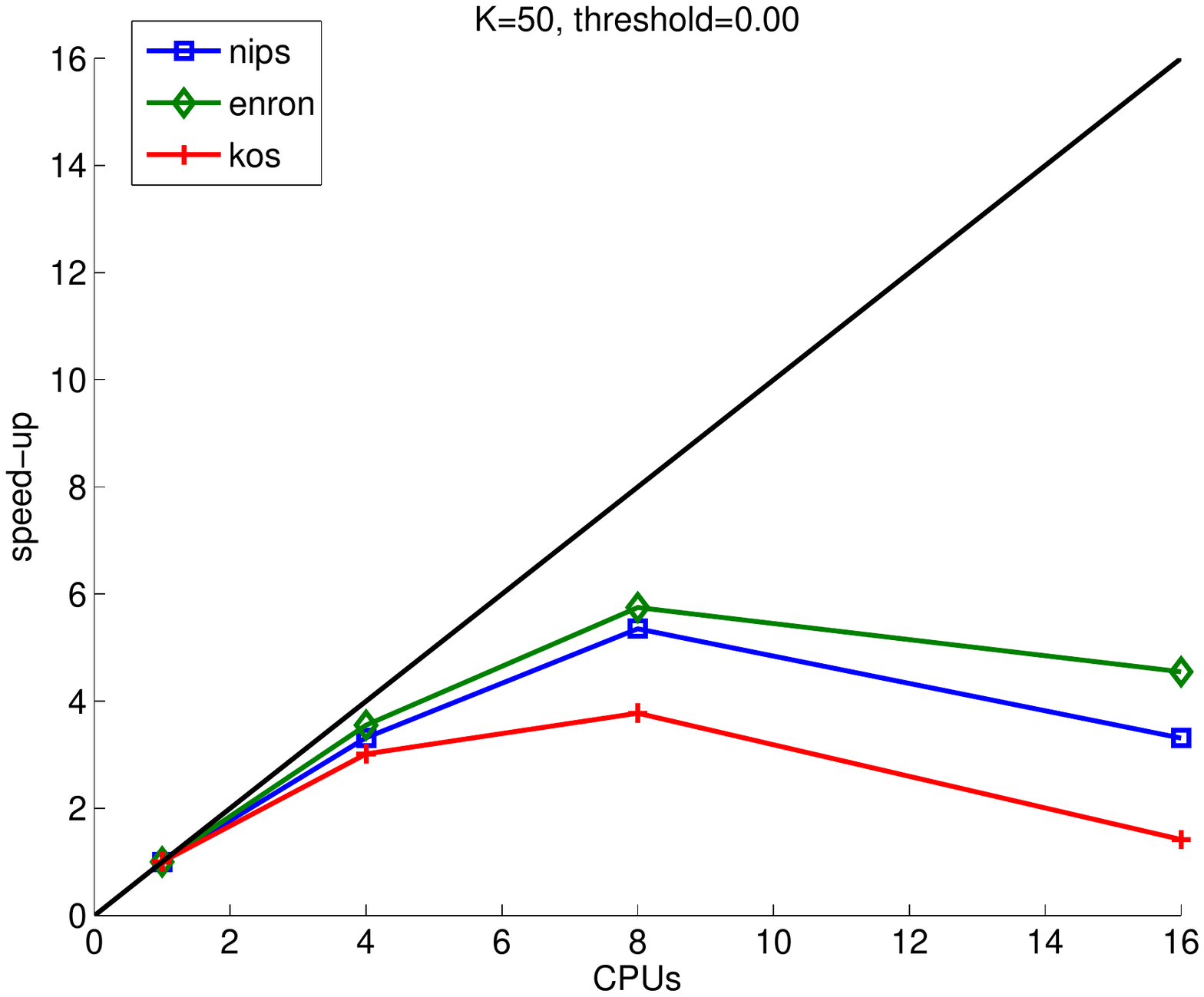}
\hspace{-10mm}
\includegraphics[width=0.6\textwidth]{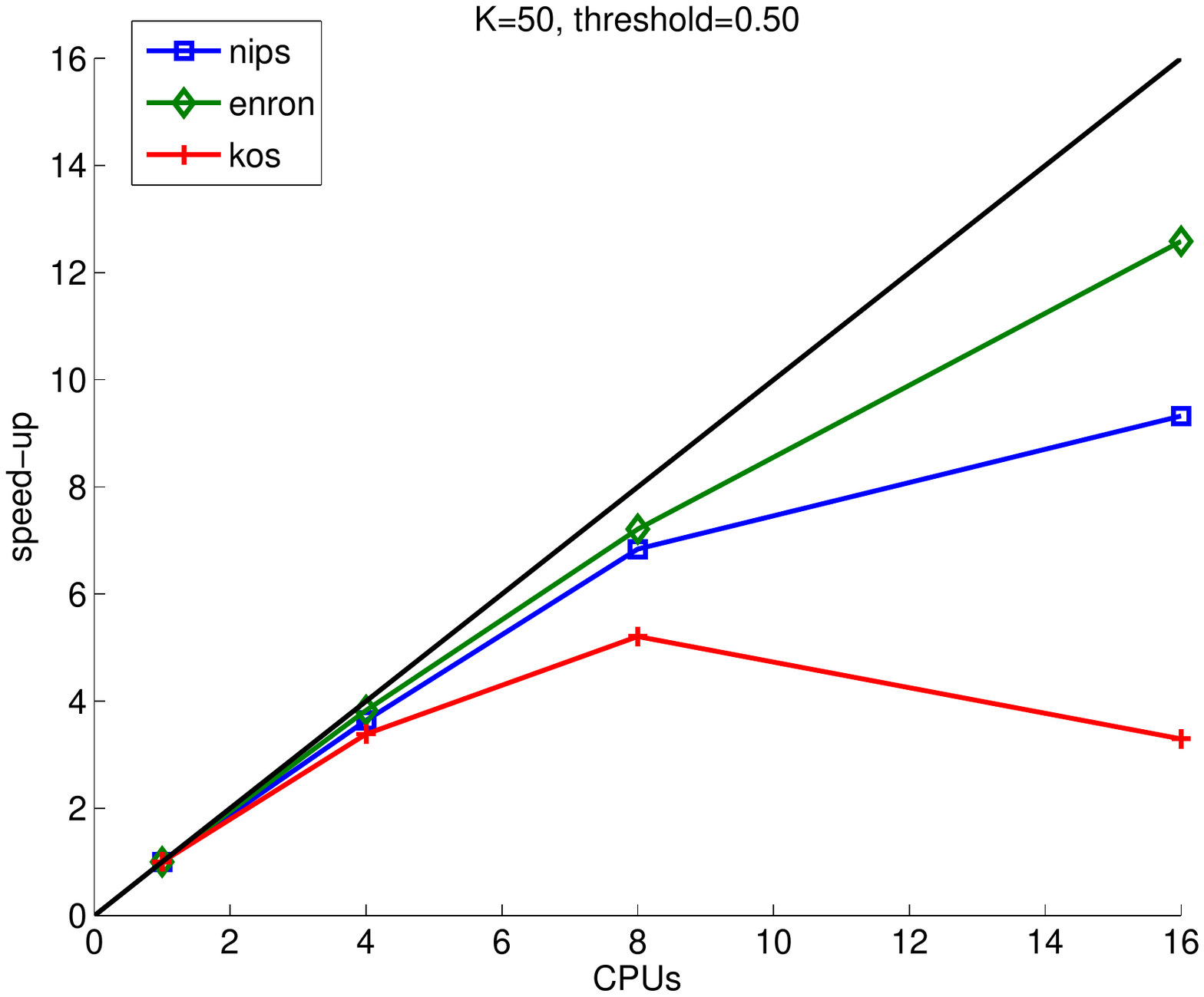}
}
\vspace{-20mm}
\caption{Speed-up for different number of CPUs ($k=50$). Left: Algorithm \ref{alg:simple} ($threshold=0$). Right: Algorithm \ref{alg:sparse} ($threshold=0.5$)}
\label{fig:speed_cpu}
\end{figure*}

In figure \ref{fig:speed_threshold} we plot the speed-up obtained for different number of CPUs with different values of $threshold$. We see that the simple sharing method (Algorithm \ref{alg:simple}), which corresponds to $threshold=0$,  fails to get a significant improvement, except for small clusters of 4 CPUs. With sparse sharing ($threshold > 0$), however, we can get speed-ups of more than 7x for 8 CPUs, and more than 12x for 16 CPUs. This can also be seen in figure \ref{fig:speed_cpu}, where we plot the speed-up for different number of CPUs for both algorithms. 

\begin{figure*}[tbp]
\centerline{
\includegraphics[width=0.6\textwidth]{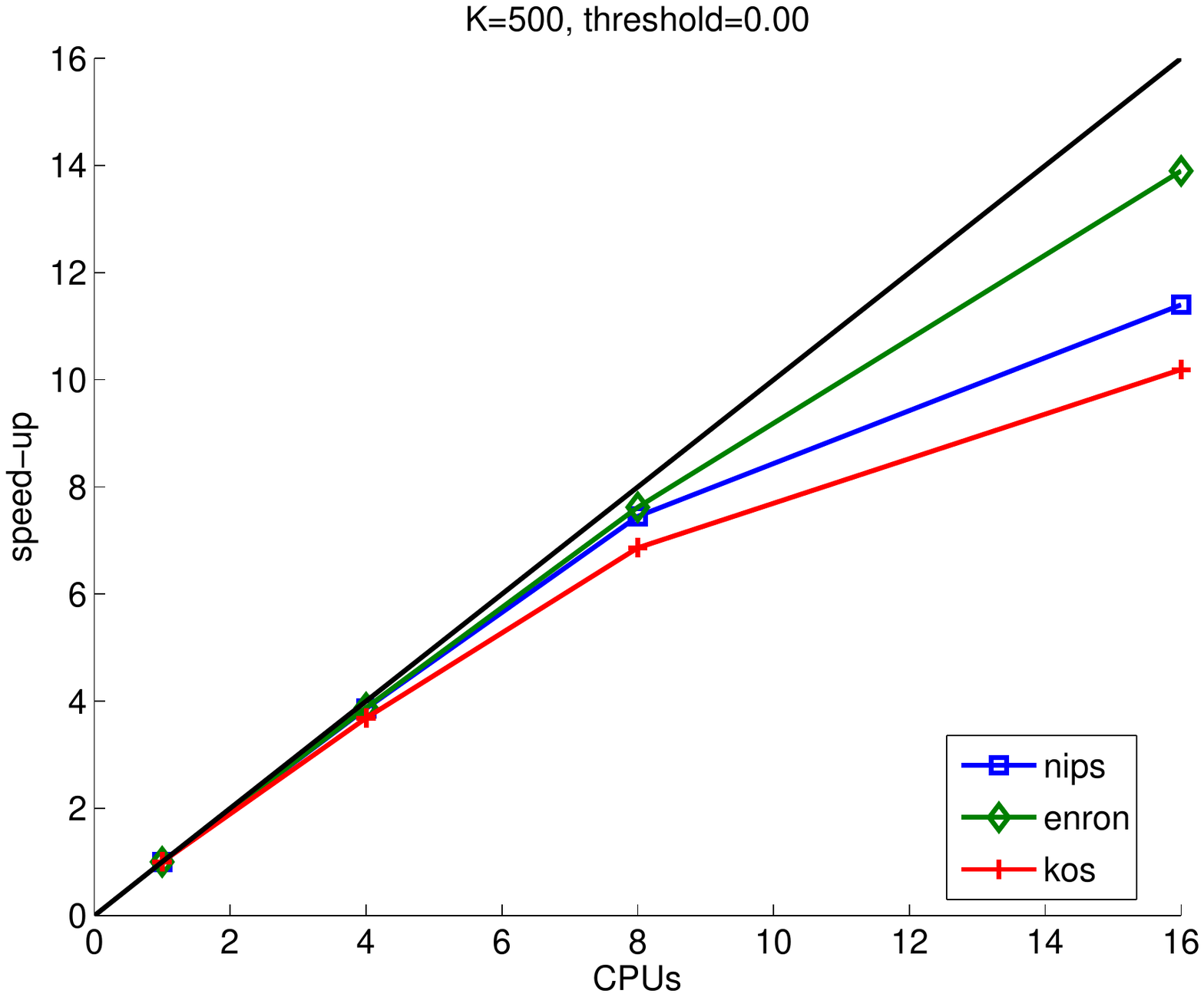}
\hspace{-10mm}
\includegraphics[width=0.6\textwidth]{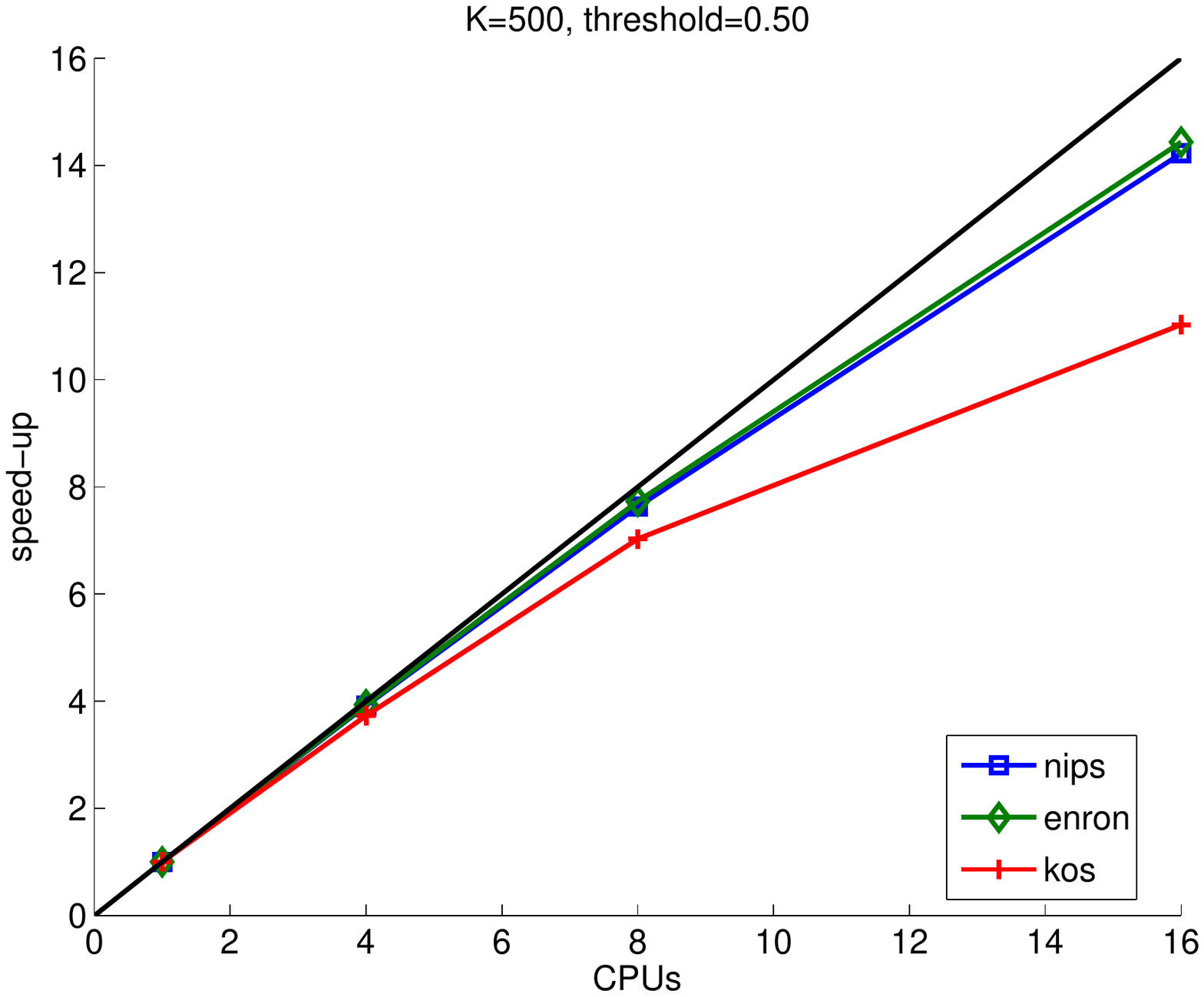}
}
\vspace{-20mm}
\caption{Speed-up for different number of CPUs ($k=500$). Left: Algorithm \ref{alg:simple} ($threshold=0$). Right: Algorithm \ref{alg:sparse} ($threshold=0.5$)}
\label{fig:speed_cpu2}
\end{figure*}

We would like to note that the datasets used are relatively small, as are the number of topics ($k=50$), leading to tasks that are not well suited for parallelization with a large number of CPUs. The purpose of these experiments was simply to measure the effects of the approximation proposed in Algorithm \ref{alg:sparse} -- for greater speed-ups when working with hundreds of CPUs a larger dataset or number of topics would be required. As an example we ran experiments with $k=500$, and as can be seen in figure \ref{fig:speed_cpu2}, we can get speed-ups closer to the theoretical limit.

\begin{figure*}[tbp]
\centerline{
\includegraphics[width=0.5\textwidth]{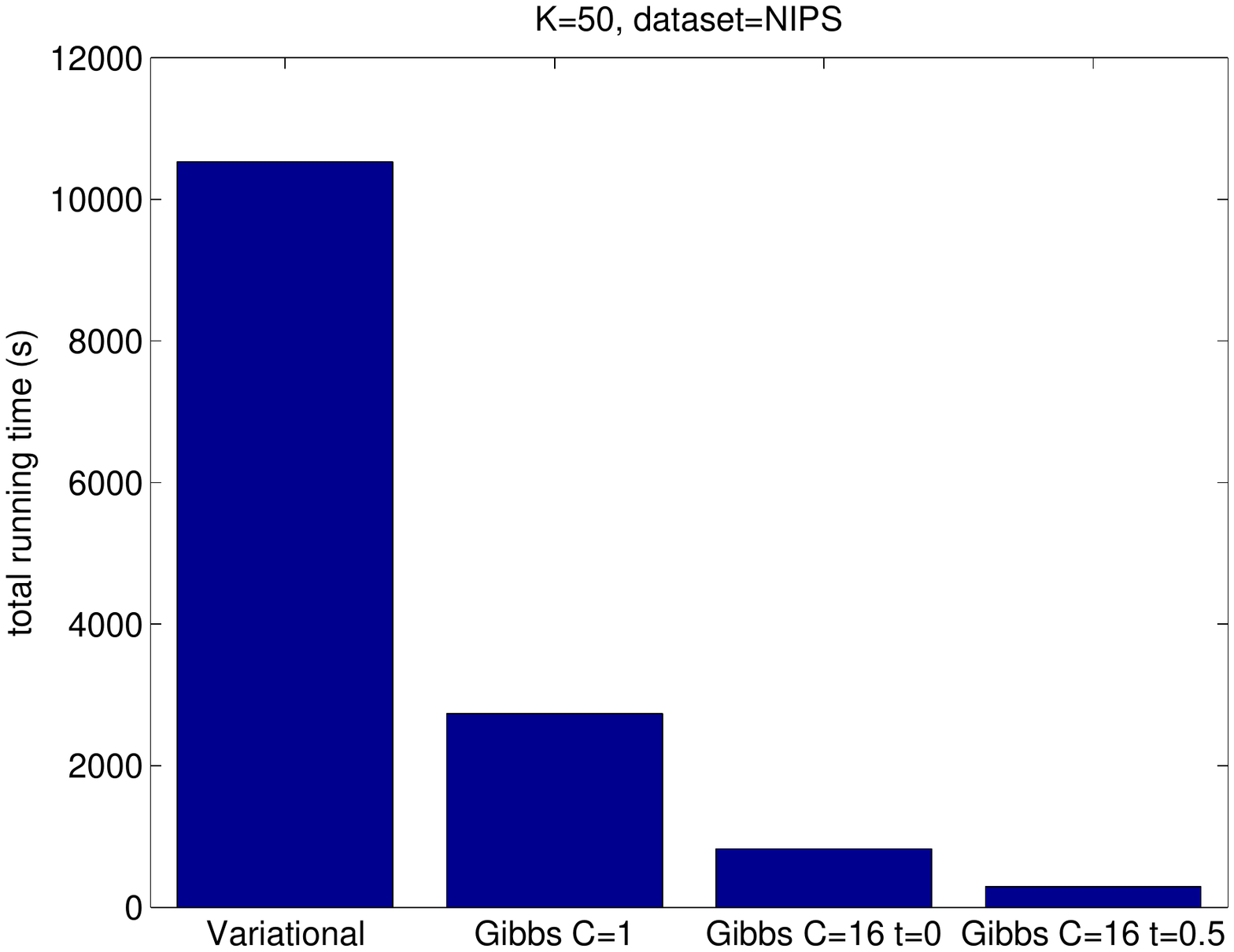}
\hspace{-10mm}
\includegraphics[width=0.5\textwidth]{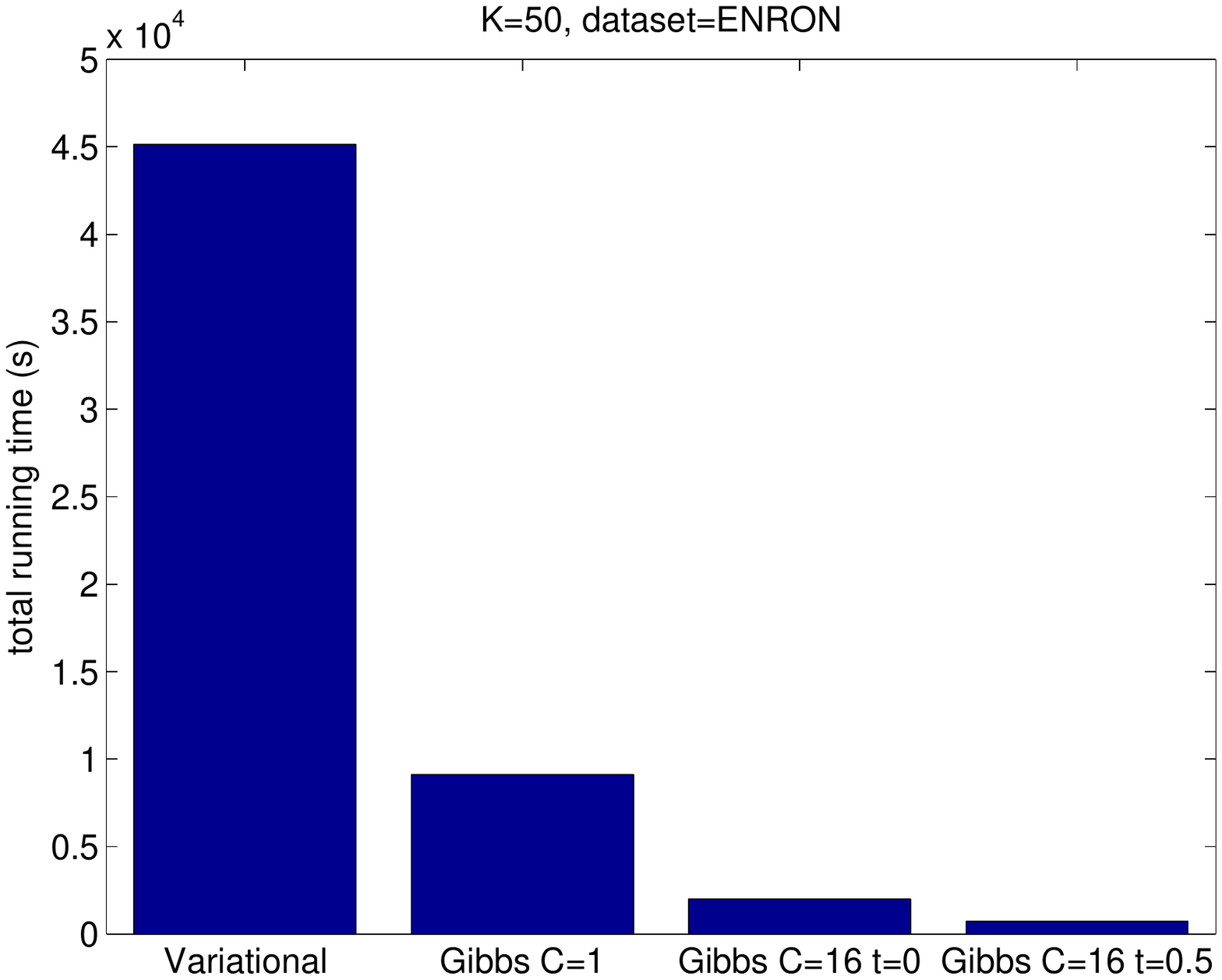}
\hspace{-10mm}
\includegraphics[width=0.5\textwidth]{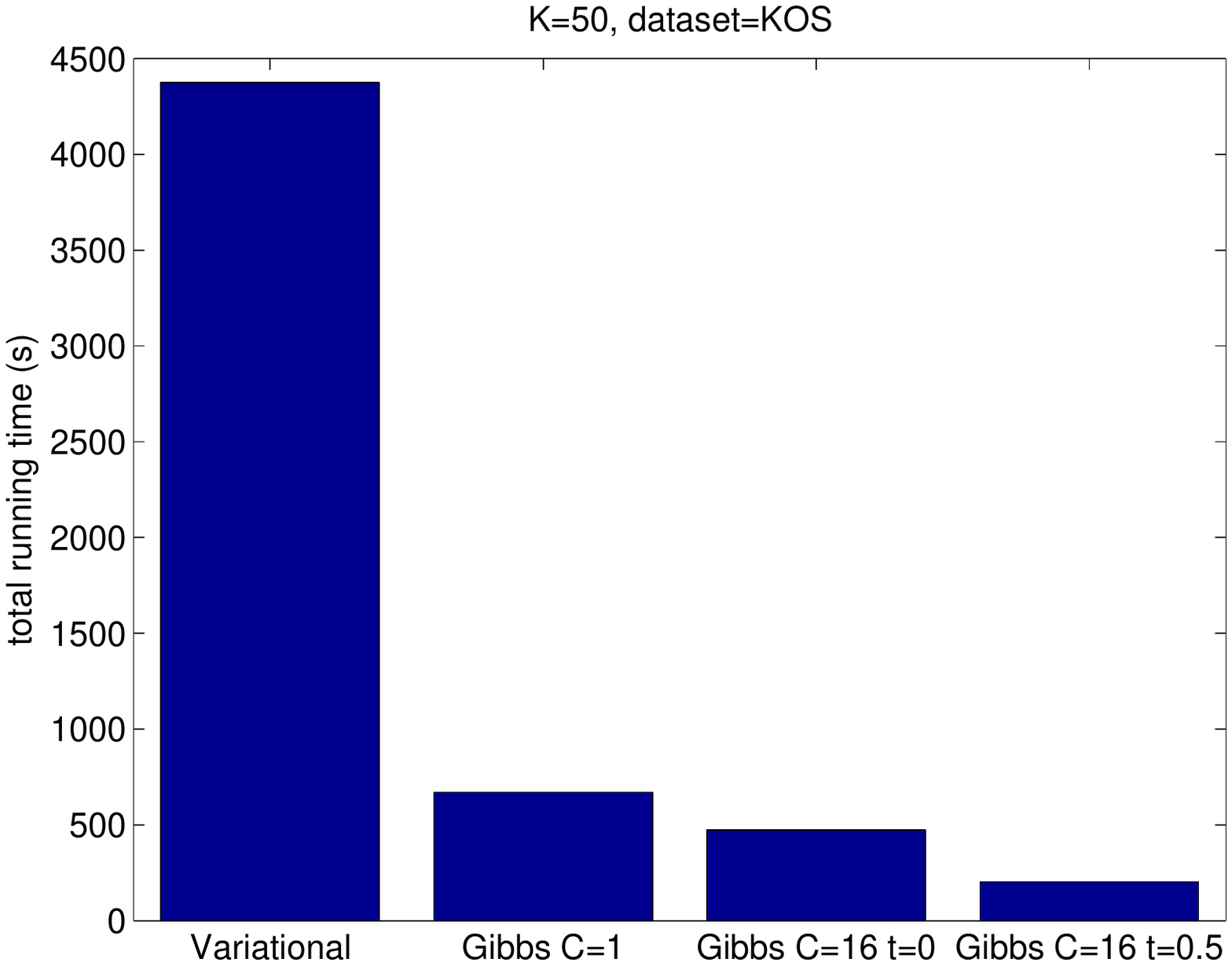}
}
\vspace{-40mm}
\centerline{
\includegraphics[width=0.5\textwidth]{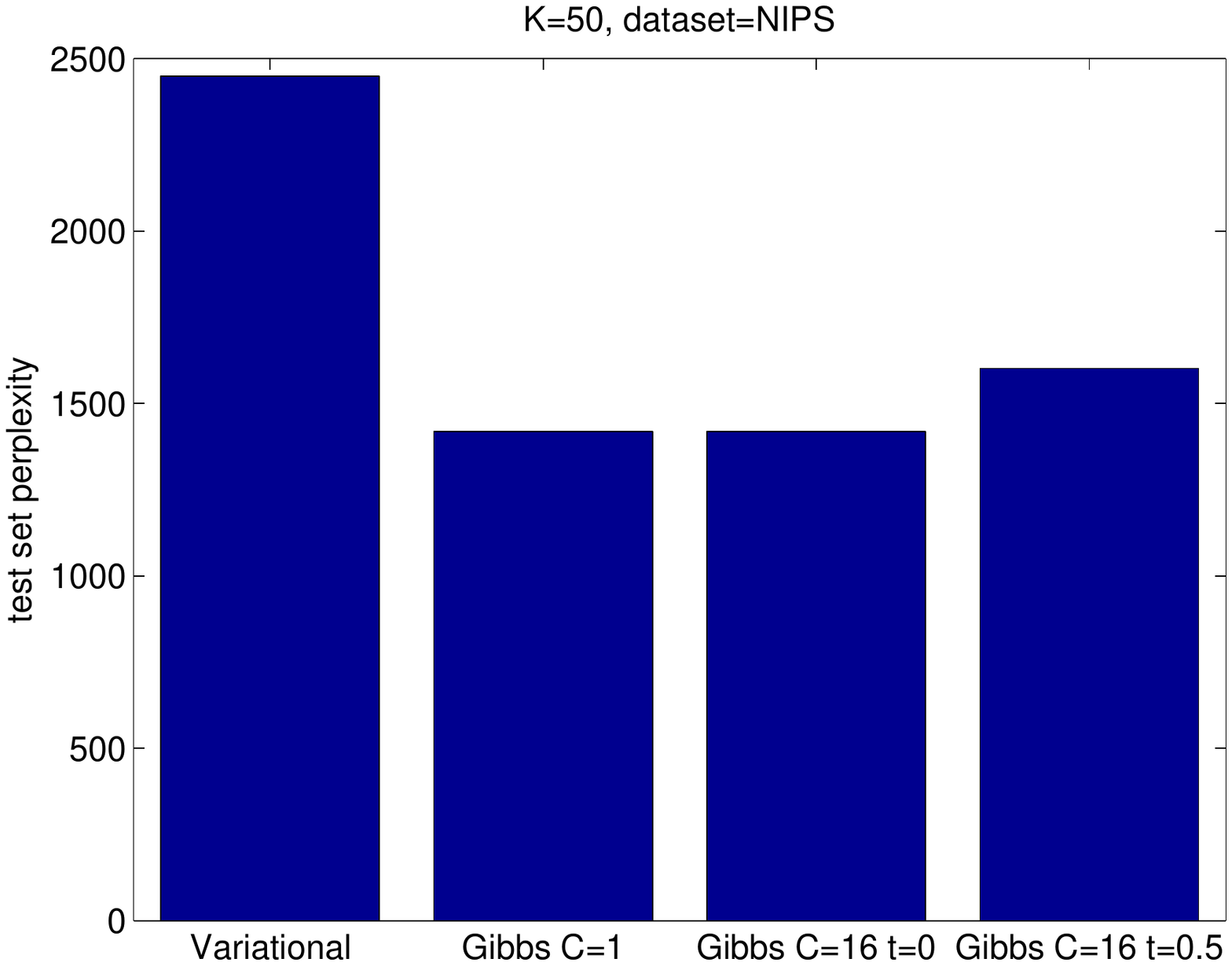}
\hspace{-10mm}
\includegraphics[width=0.5\textwidth]{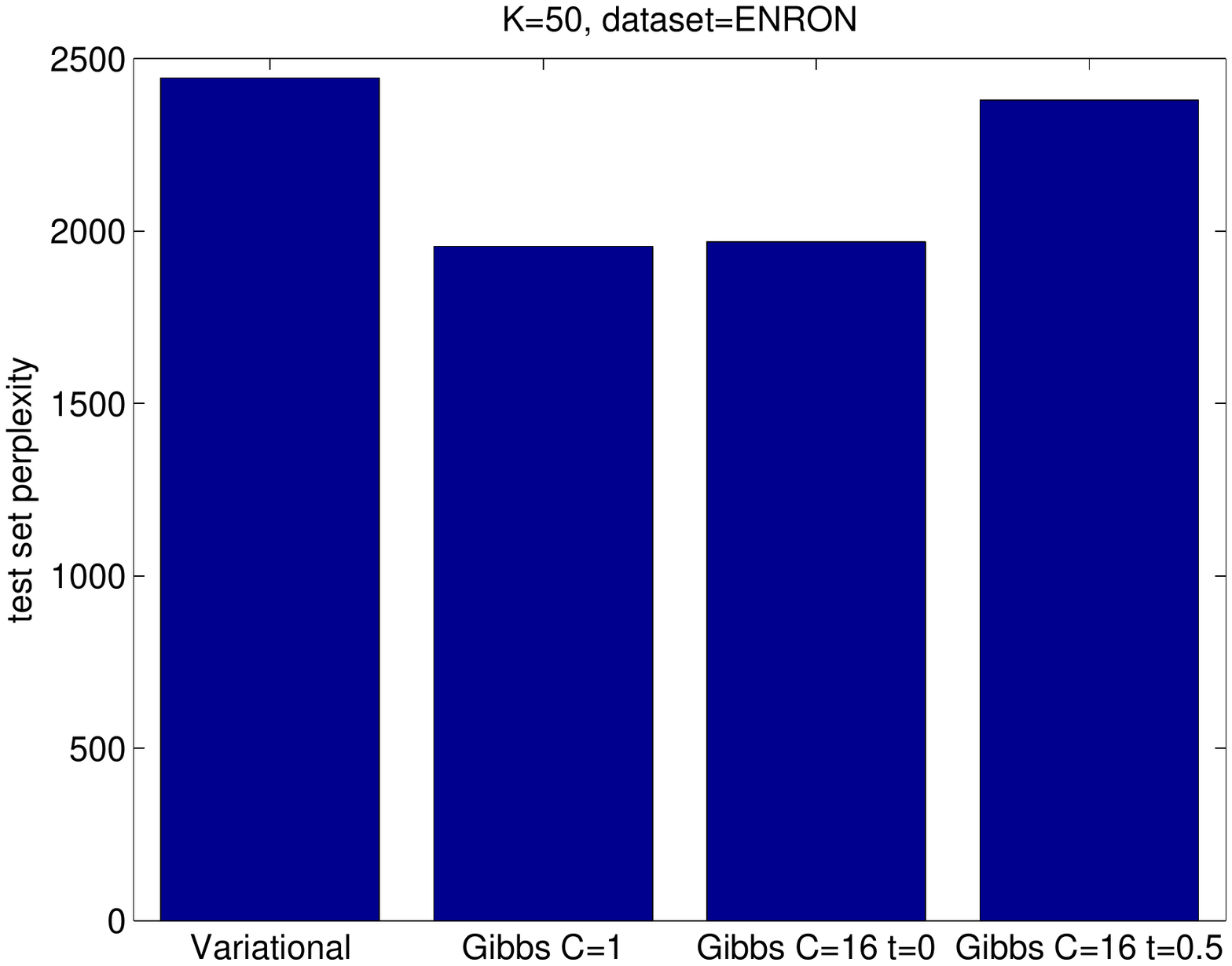}
\hspace{-10mm}
\includegraphics[width=0.5\textwidth]{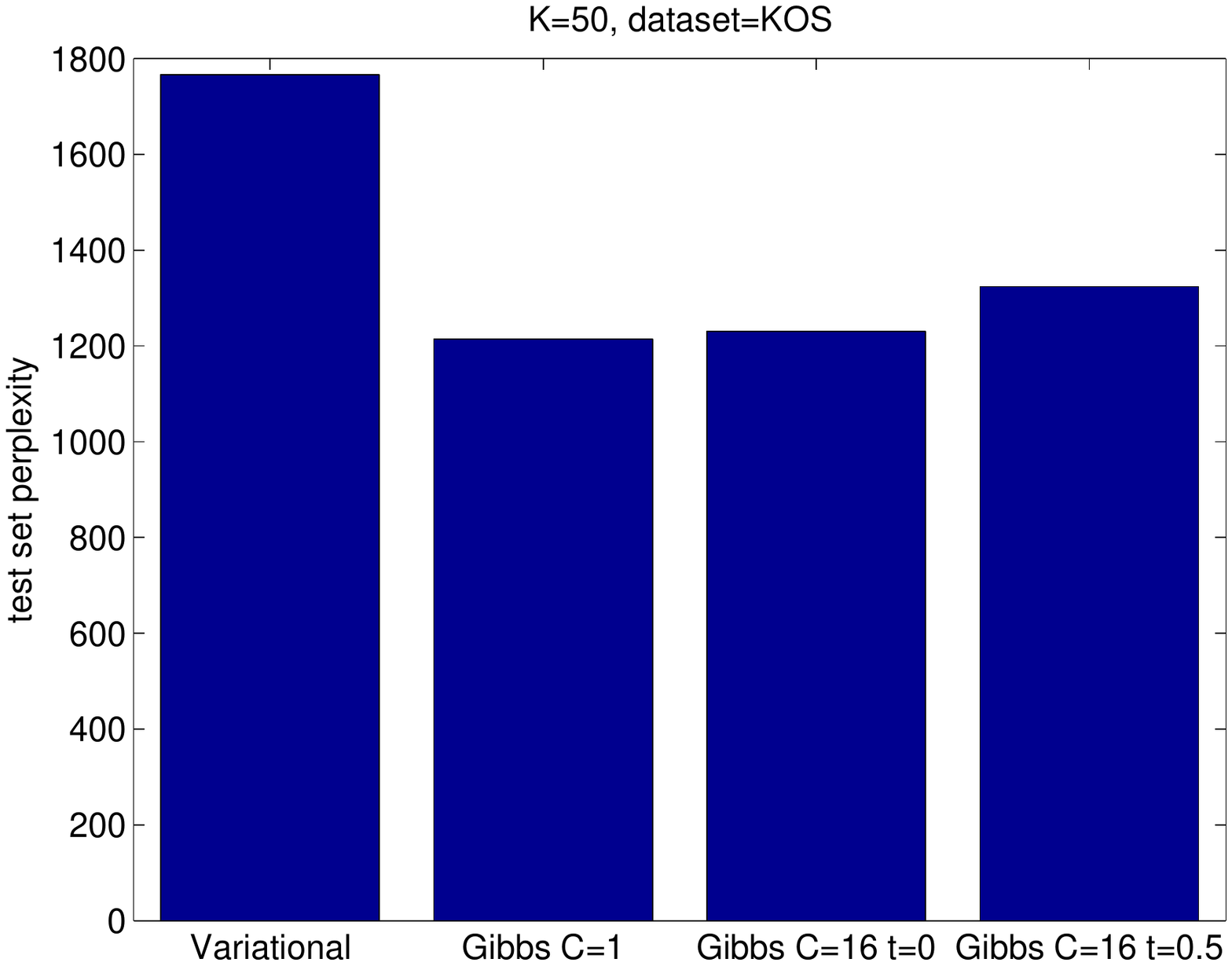}
}
\vspace{-20mm}
\caption{Comparison to variational Bayesian inference. Top: running time. Bottom: test set perplexity. From left to right: NIPS, ENRON and KOS datasets. $C$ is the number of CPUs, $t$ is the $threshold$ parameter.}
\label{fig:comp_varational}
\end{figure*}

To get some perspective on the significance of the approximations being used, in figure \ref{fig:comp_varational} we compare our results to a variational Bayes inference implementation. We used the code from \cite{BleNgJor03}\footnote{\href{http://www.cs.princeton.edu/~blei/lda-c/index.html}{http://www.cs.princeton.edu/$\sim$blei/lda-c/index.html}}, with its default parameters, and $\alpha$ fixed to $0.1$, as in the Gibbs experiments. As can be seen, not only the Gibbs sampler is substantially faster, its perplexity results are better, even with all the approximations.

\section{Conclusion and Discussion}

We proposed a simple method to reduce the amount of time spent in synchronization in a distributed implementation of LDA. We present empirical results showing a reasonable speed-up improvement, at the cost of a small reduction in the quality of the learned model. The method is tunable, allowing a trade off between speed and accuracy, and is completely asynchronous. Source code is available at the first authors' web page.\footnote{\href{http://users.rsise.anu.edu.au/~jpetterson/}{http://users.rsise.anu.edu.au/$\sim$jpetterson/}}

As future work we plan to look for more efficient ways of sharing information among CPUs, while also applying the method to larger datasets, where we expect to see more significative speed-up improvements.

\bibliographystyle{alpha}
\newcommand{\etalchar}[1]{$^{#1}$}

\end{document}